\def\year{2020}\relax

\documentclass[letterpaper]{article} % DO NOT CHANGE THIS
\usepackage{aaai20}  % DO NOT CHANGE THIS
\usepackage{times}  % DO NOT CHANGE THIS
\usepackage{helvet} % DO NOT CHANGE THIS
\usepackage{courier}  % DO NOT CHANGE THIS
\usepackage[hyphens]{url}  % DO NOT CHANGE THIS
\usepackage{graphicx} % DO NOT CHANGE THIS
\def\UrlFont{\rm}  % DO NOT CHANGE THIS
\usepackage{graphicx}  % DO NOT CHANGE THIS
\usepackage{cdann_definitions}

\newcommand{\cvar}{\operatorname{CVaR}}
\allowdisplaybreaks

\newcommand{\alex}[1]{\textcolor[rgb]{.9,0,0}{}}%\bf [Alex: #1]}}

\newcommand{\Vmin}{V_{\min}}
\newcommand{\Vmax}{V_{\max}}
\newcommand{\shift}{c}

\usepackage[ruled, vlined, linesnumbered]{algorithm2e}

\SetCommentSty{mycommfont}
\usepackage{booktabs}

\usepackage{tikz}

\newcommand*\circledmarker[1]{\tikz[baseline=(char.base)]{
            \node[shape=circle,draw,inner sep=.5pt] (char) {\scriptsize #1};}}
\newcommand*\circledmarked[2]{\overset{\circledmarker{#1}}{#2}}

\title{Being Optimistic to Be Conservative: Quickly Learning a CVaR Policy}
\author{
Ramtin Keramati\textsuperscript{\rm 1}, Alex Tamkin,\textsuperscript{\rm 2} Christoph Dann\textsuperscript{\rm 3}, Emma Brunskill\textsuperscript{\rm 2}\\
\textsuperscript{\rm 1}Institute of Computational and Mathematical Engineering (ICME), Stanford University, California, USA\\ 
\textsuperscript{\rm 2}Department of Computer Science, Stanford University, California, USA\\
\textsuperscript{\rm 3} Machine Learning Department, Carnegie Mellon University, Pittsburgh, Pennsylvania, USA\\
\texttt{\{keramati,atamkin,ebrun\}@cs.stanford.edu}, \texttt{cdann@cdann.net}
}
 
\begin{document}

\maketitle

\begin{abstract}
While maximizing expected return is the goal in most reinforcement learning approaches, risk-sensitive objectives such as conditional value at risk (CVaR) are more suitable for many high-stakes applications. However, relatively little is known about how to explore to quickly learn policies with good CVaR. In this paper, we present the first algorithm for sample-efficient learning of CVaR-optimal policies in Markov decision processes based on the optimism in the face of uncertainty principle. This method relies on a novel optimistic version of the distributional Bellman operator that moves probability mass from the lower to the upper tail of the return distribution. We prove asymptotic convergence and optimism of this operator for the tabular policy evaluation case. We further demonstrate that our algorithm finds CVaR-optimal policies substantially faster than existing baselines in several simulated environments with discrete and continuous state spaces.
\end{abstract}

\section{Introduction}
\label{sec:intro}
%\chris{paragraph on why risk-sensitive policies in RL}

A key goal in reinforcement learning (RL) is to quickly learn to make good decisions by interacting with an environment. In most cases the quality of the decision policy is evaluated with respect to its  expected (discounted) sum of rewards. However, in many interesting cases, it is important to consider the full distributions over the potential sum of rewards, and the desired objective may be a risk-sensitive measure of this distribution. For example, a patient undergoing a surgery for a knee replacement will (hopefully) only experience that procedure once or twice, and may will be interested in the distribution of potential results for a single procedure, rather than what may happen on average if he or she were to undertake that procedure hundreds of time. Finance and (machine) control are other cases where interest in risk-sensitive outcomes are common.

%However, this objective fails to address phenomena like loss aversion, where the subjective  impact of a loss is often greater than that of an equivalent gain \citep{tversky1992advances}.  This has particular relevance in high-stakes settings. For example, in finance, people may  prefer investment strategies which yield modest but stable returns over higher-yield strategies  that have a chance of bankrupting them. Likewise, patients undergoing surgery might opt for a procedure with a longer recovery time if it means minimizing the already-small chance of a serious complication.

A popular risk-sensitive measure of a distribution of outcomes is the Conditional Value at Risk (CVaR) \cite{artzner1999coherent}. Intuitively, CVaR is the expected reward in the worst $\alpha$-fraction of outcomes, and has seen extensive use in financial portfolio optimization 
\cite{finance-zhu2009worst}, often under the name ``expected shortfall''. While there has been recent interest in the RL community in learning to converge or identify good CVaR decision policies 
in Markov decision processes ~\cite{chow2014algorithms,chow2015risk,sampling,dabney2018implicit}, interestingly we are unaware of prior work focused on how to quickly learn such CVaR MDP policies, even though sample efficient RL for maximizing expected outcomes is a deep and well studied theoretical~\cite{Jaksch10,dann2018policy} and empirical \cite{bellemare2016unifying} topic. Sample efficient exploration seems of equal or even more importance in the case when the goal is risk-averse outcomes. 

In this paper we work towards sample efficient reinforcement learning algorithms that can quickly identify a policy with an optimal CVaR. Our focus is in minimizing the amount of experience needed to find such a policy, similar in spirit to probably approximately correct RL methods for expected reward. Note that this is different than another important topic in risk-sensitive RL, which focuses on safe exploration: algorithms that focus on avoiding any potentially very poor outcomes during learning. These  typically rely on local smoothness assumptions and do not typically focus on sample efficiency~\cite{berkenkamp2017safe,koller2018learning}; an interesting question for future work is whether one can do both safe and efficient learning of a CVaR policy. Our work is suitable for the many settings where some outcomes are undesirable but not catastrophic.

Our approach is inspired by the popular and effective principle of  optimism in the face of uncertainty (OFU) in sample efficient RL for maximizing expected outcomes~\cite{strehl2008analysis,brafman2002r}. Such work typically works by considering uncertainty over the MDP model parameters or state-action value function, and constructing an optimistic value function given that uncertainty that is then used to guide decision making. To take a similar idea for rapidly learning the optimal CVaR policy, we seek to consider the uncertainty in the distribution of outcomes possible and the resulting CVaR value. To do so, we use the Dvoretzky-Kiefer-Wolfowitz (DKW) inequality---while to our knowledge this has not been previously used in reinforcement learning settings, it is a very useful concentration inequality for our purposes as it provides bounds on the true cumulative distribution function (CDF) given a set of sampled outcomes. We leverage these bounds in order to compute optimistic estimates of the optimal CVaR. 

Our interest is in creating empirically efficient and scalable algorithms that have a theoretically sound grounding. To that end, we introduce a new algorithm for quickly learning a CVaR policy in MDPs and show that at least in the evaluation case in tabular MDPs, this algorithm indeed produces optimistic estimates of the CVaR. We also show that it does converge eventually. We accompany the theoretical evidence with an empirical evaluation. We provide encouraging empirical results on a machine replacement task~\cite{delage2010percentile}, a classic MDP where risk sensitive policies are critical, as well as a well validated simulator for type 1 diabetes~\cite{man2014uva} and a simulated treatment optimization task for HIV~\cite{ernst2006clinical}. In all cases we find a substantial benefit over simpler exploration strategies. To our knowledge this is the first algorithm that performs strategic exploration to learn good CVaR MDP policies.

\section{Background and Notation}
\label{sec:background}

Let $X$ be a bounded random variable with cumulative distribution function $F(x)=\mathbb{P}[X\leq x]$. The 
\emph{conditional value at risk (CVaR)} at level $\alpha \in (0,1)$ of a random variable $X$ is then defined as~\cite{rockafellar2000optimization}:
\begin{align}
    \cvar_\alpha(X) := \sup_\nu \left\{\nu - \frac{1}{\alpha}\EE[{(\nu - X)^+]}\right\}
\end{align}
We define the inverse CDF as $F^{-1}(u) = \inf\{ x : F(x) \geq u\}$. It is well known that when $X$ has a continuous distribution, $\cvar_\alpha(X) = \EE_{X \sim F}[X | X \leq F^{-1}(\alpha)]$ \cite{acerbi2002coherence}. For ease of notation we sometimes write $\cvar$ as a function of the CDF $F$, $\cvar_\alpha(F)$. 

We are interested in the $\cvar$ of the discounted cumulative reward in a \textbf{Markov Decision Process (MDP)}. An MDP is defined by a tuple $(\Scal, \Acal, R, P, \gamma)$, where $\Scal$ and $\Acal$ are finite state and action space, $r \sim R(s,a)$ is the reward distribution, $s' \sim P(s,a)$ is the transition kernel and $\gamma \in [0,1)$ is the discount factor. A stationary policy $\pi$ maps each state $s \in \Scal$ to a probability distribution over action space $\Acal$. 

Let $\Zcal$ denote the space of distributions over returns (discounted cumulative rewards) from such an MDP, and assume that these returns are in $[\Vmin, \Vmax]$ almost surely, where $\Vmin \geq 0$. We define $Z_{\pi}(s,a) \in \Zcal$ to be the distribution of the return of policy $\pi$  with CDF $F_{Z_{\pi}(s,a)}$ and initial state action pair $(s,a) \in \Scal \times \Acal$ as 
$Z_{\pi}(s,a) := \text{Law}_\pi\left(\sum_{t=0}^\infty \gamma^t R_t|S_0=s, A_0=a\right)
$. RL algorithms most commonly optimize policies for expected return and explicitly learn Q-values, $Q^\pi(s,a) = \EE[Z_\pi(s,a)]$ by applying approximate versions of Bellman backups. Instead, we are interested in other properties of the return distribution and we will build on several recently proposed algorithms that aim to learn a parametric model of the entire return distribution instead of only its expectation. Such approaches are known as \emph{distributional RL methods}.

\paragraph{Distributional Reinforcement Learning}
Distributional RL methods apply a sample-based approximation to distributional versions of the usual Bellman operators. For example, one can define a distributional Bellman operator \cite{bellemare2017distributional} as $
\Tcal^{\pi}: \Zcal \rightarrow \Zcal$ as 
\begin{align}\label{eq:DRL}
    \Tcal^\pi Z_{\pi}(s,a) \stackrel{D}{:=} R(s,a) + \gamma P^{\pi}Z(s,a)
\end{align}
where $\stackrel{D}{=}$ denotes equality in distribution, and the transition operator is defined as $P^{\pi}Z(s,a) \stackrel{D}{:=} Z(s',a')$ with $s' \sim P(\cdot|s,a)$, $a' \sim \pi(s)$. The optimality version $\Tcal$ is similarly any $\Tcal Z = \Tcal^\pi Z$ where $\pi$ is an optimal policy w.r.t. expected return. Note that this is not necessarily unique when there are multiple optimal policies. \cite{rowland2018analysis} showed that $\Tcal^\pi$ is a $\sqrt{\gamma}$-contraction in the Cram\'er-metric, $\bar \ell_2$
\begin{align}
    \bar \ell_2(Z_1, Z_2) = & \,\sup_{s,a} \ell_2(Z_1(s,a), Z_2(s,a))  \\
    =& \, \sup_{s,a} \left(\int (F_{Z_1(s,a)}(u) - F_{Z_2(s,a)}(u))^2 du\right)^{1/2}\nonumber
\end{align}
One of the canonical algorithms in distributional RL is 
CDRL or C51 \cite{bellemare2017distributional} which represent the return distribution $Z^\pi$ as a discrete distribution with fixed support on $N$ atoms $\{z_i = V_{\min} + i \Delta z: 0 \leq i < N \}, \Delta z := \frac{V_{\max}-V_{\min}}{N-1}$ the discrete distribution is parameterized as $\theta:\Scal \times \Acal \rightarrow \mathbb{R}^N$:
\begin{align*}
    Z_{\theta}(s,a) = z_i \quad \text{w.p.} \quad p_i(s,a)=\frac{e^{\theta_i(s,a)}}{\sum_j e^{\theta_j(s,a)}}.
\end{align*}
Essentially, C51 uses a sample transition $(s,a,r,s')$ to perform an approximate Bellman backup $Z \gets \Pi_{\Ccal} \hat \Tcal Z$, where $\hat \Tcal$ is a sample-based Bellman operator and $\Pi_{\Ccal}$ is a projection back onto the support of discrete distribution $\{z_{0}, \dots, z_{N-1}\}$.

\section{Optimistic Distributional Operator}

\begin{figure}
    \centering
    \includegraphics[width=0.9\linewidth]{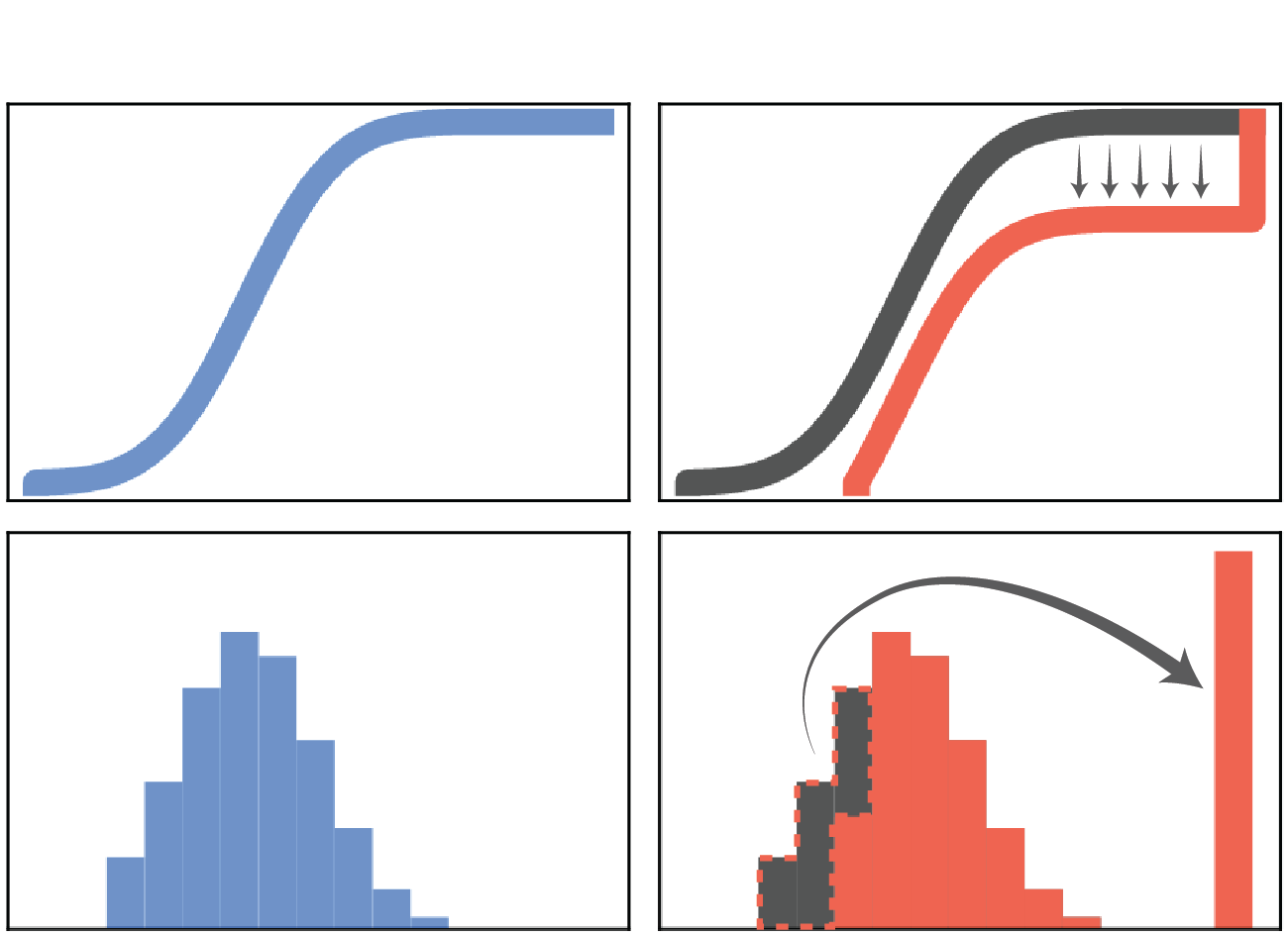}
    \caption{Top-left: Empirical CDF Top-right: The lower DKW confidence band (a shifted-down version of the empirical CDF). Bottom-left: Empirical PDF. Bottom-right: Optimistic PDF.}
    \label{fig:dkw_cartoon}
\end{figure}

In contrast to the typical RL setup where an agent tries to maximize its expected return, we seek to learn a stationary policy that maximizes the $\cvar_{\alpha}$ of the return at risk level $\alpha$.\footnote{Note that the $\cvar$-optimal policy at any state can be non-stationary~\cite{shapiro2009lectures}, as it depends on the sum of rewards achieved up to that state. For simplicity, as \cite{dabney2018distributional} we instead seek a stationary policy, which will generally can be suboptimal but typically still achieve high CVaR, as observed in our experiments.} To find such policies quickly, we follow the optimism-in-the-face-of-uncertainty (OFU) principle and introduce optimism in our CVaR estimates to guide exploration. 
While adding a bonus to rewards is a popular approach for optimism in the standard expected return case \cite{ostrovski2017count}, we here follow a different approach and introduce optimism into our return estimates by shifting the empirical CDFs. Formally, consider a return distribution $Z(s,a) \in \Zcal$ with CDF $F_{Z(s,a)}(x)$. We define the optimism operator $O_\shift: \Zcal \rightarrow \Zcal$ as
\begin{align}\label{eq:opt_op}
    &F_{O_\shift Z(s,a)}(x) = \left( F_{Z(s,a)}(x) - \shift\frac{ \one\{x \in [\Vmin, \Vmax)\}}{\sqrt{n(s,a)}}\right)^{+}
\end{align}
where $\shift$ is a constant and $(\cdot)^+$ is short for $\max\{\cdot, 0\}$. In the definition above, $n(s,a)$ is the number of times the pair $(s,a)$ has been observed so far or an approximation such as pseudo-counts \cite{bellemare2016unifying}. By shifting the cumulative distribution function down, this operator essentially puts probability mass from the lower tail to the highest possible value $\Vmax$. An illustration is provided in Figure~\ref{fig:dkw_cartoon}.
This approach to optimism is motivated by an application of the DKW-inequality to the empirical CDF. As shown recently by \cite{thomas2019concentration}, this can yield tighter upper confidence bounds on the CVaR. %Our approach also has the advantage that the scaling $\shift$ does not depend on the risk level $\alpha$, unlike a potential reward bonus approach that would likely require more parameter tuning. % EB: we say this again in theory, and also we don't compare to potential reward bonus approach
\section{Theoretical Analysis}
The optimistic operator introduced above operates on the entire return distribution and our algorithm introduced in the next section combines this optimistic operator to estimated return-to-go distributions. As such, it belongs to the family of distributional RL methods~\cite{dabney2018distributional}. These methods are a recent development and come with strong asymptotic convergence guarantees when used for \emph{policy evaluation} in tabular MDPs~\cite{rowland2018analysis}. Yet, finite sample guarantees such as regret or PAC bounds still remain elusive for distributional RL \textit{policy optimization} algorithms. 

A key technical challenge in proving performing bounds for distributionally robust policy optimization during RL is that convergence of the distributional Bellman optimality operator can generally not be guaranteed. Prior results have only showed that if the optimization process itself is to compute a policy which maximizes expected returns, such as Q-learning, then convergence of the distirbutional Bellman optimality operator is guaranteed to converge.~\cite[Theorem 2]{rowland2018analysis}. Note however that if the goal is to leverage distributional information to compute a policy to maximize something other than expected outcomes, such as a risk sensitive policy like we consider here, no prior theoretical results are known in the reinforcement learning setting to our knowledge. However, it is promising that there is some empirical evidence that one can compute risk-sensitive policies using 
distributional Bellman operators~\cite{dabney2018implicit} which suggests that more theoretical results may be possible. 

Here we take a first step towards this goal.  
%Fortunately, convergence issues eof the distributional Bellman optimality have not been observed empirically which suggests that better theoretical characterizations could enable finite-sample guarantees eventually. Developing such characterizations is an important problem but not the focus of this paper. 
Our primary aim in this work is to provide tools to introduce optimism into distributional return-to-go estimates to guide sample-efficient exploration for CVaR. Therefore, our theoretical analysis focuses  on showing that this form of optimism does not harm convergence and is indeed a principled way to obtain optimistic CVaR estimates. 

%Theoretical properties of distributional RL are not yet well understood. Recent advances provide convergence guarantees for tabular distributional RL under the Bellman optimality operator; however, convergence with function approximation remains an open problem.~\cite{dabney2018distributional} These results leverages the convergence of Q-learning~\cite{watkins1992q} algorithm, hence additional challenges arise when considering distorted expectations (e.g. $\cvar$) which makes the convergence of distributional RL under distorted expectations yet an open problem, even in tabular setting. 

%The optimism operator $O_\shift$ in the previous section is motivated by the Dvoretzky-Kiefer-Wolfowitz inequality and has some desirable properties. Specifically, similar to \cite{bellemare2017distributional} we show this shift in CDF will lead to converging behavior when used in a distributional Bellman backup and this indeed yields an optimistic CVaR estimates.   

First, we prove that the optimism operator is a non-expansion in the Cram\'er distance. This results shows that this operator can be used with other contraction operators without negatively impacting the convergence behaviour. Specifically we can guarantee convergence with distributional Bellman backup.
\begin{proposition}\label{mdp:contraction}
    For any $\shift$, the $O_\shift$ operator is a non-expansion in the Cram\'er distance $\bar\ell_2$. This implies that optimistic distributional Bellman backups $O_\shift \Tcal^\pi$ and the projected version $\Pi_{\Ccal} O_\shift \Tcal^\pi$ are $\sqrt{\gamma}$-contractions in $\bar\ell_2$ and iterates of these operators converge in $\bar\ell_2$ to a unique fixed-point.
\end{proposition}

% This result does not include Bellman backups with the Bellman-optimality operator $\Tcal$ or its generalization $\Tcal_\alpha$ to CVaR-greedy policies. Here, $\Tcal_{\alpha}$ is any operator  that satisfies
%  $\Tcal_{\alpha} Z = \Tcal^\pi Z$ for any policy $\pi$ that is greedy w.r.t. CVaR at level $\alpha$, i.e., $\pi(s) \in \argmax_{a} \cvar_\alpha(Z(s,a))$. Note that $\Tcal_{\alpha} = \Tcal$ for $\alpha = 1$. The lack of general convergence guarantees in the control case is because optimality operators are generally not contractions \cite{bellemare2017distributional} and not a limitation of our optimism operator and we have not observed any convergence issues in our experiments.
 
Next, we provide theoretical evidence that this operator indeed produces optimistic CVaR estimates. Consider here batch policy evaluation in MDPs $M$ with finite state- and action-spaces. Assume that we have collected a fixed number of samples $n(s,a)$ (which can vary across states and actions) and build an empirical model $\hat M$ of the MDP. For any policy $\pi$, let $\hat \Tcal^\pi$ denote the distributional Bellman operator in this empirical MDP. Then we indeed achieve optimistic estimates by the following result:

\begin{theorem}\label{mdp:optimistic_cvar}
Let the shift parameter in the optimistic operator be sufficiently large which is $\shift = O\left( \ln ( |\Scal||\Acal| / \delta)\right)$. Then with probability at least $1 - \delta$, the iterates $\cvar_{\alpha}((O_{\shift} \hat \Tcal^\pi)^m Z_0)$ converges for any risk level $\alpha$ and initial $Z_0 \in \Zcal$ to an optimistic estimate of the policy's conditional value at risk. That is, with probability at least $1 - \delta$, 
\begin{align*}
    \forall s,a : \cvar_{\alpha}((O_{\shift} \hat \Tcal^\pi)^\infty Z_0(s,a)) \geq \cvar_{\alpha}(Z_\pi(s,a)).
\end{align*}
\end{theorem}

This theorem uses the DKW inequality which to the best of our knowledge has not been used for MDPs.
Note, that the statement guarantees optimism for all risk levels $\alpha \in [0,1]$ without paying a penalty for it. Since we estimate the transitions and rewards for each state and action separately, one generally does not expect to be able to use a shift parameter smaller than $\Omega(\ln(|\Scal| |\Acal| /\delta))$. Thus, Theorem~\ref{mdp:optimistic_cvar} is unimprovable in that sense.
Specifically, we avoid a polynomial dependency on the number of states $|\Scal|$ in the shift parameter $c$ by combining two techniques: (1) concentration inequalities w.r.t. the optimal CVaR of the next state for a certain finite set of alphas and (2) a covering argument to get optimism for all infinitely many $\alpha \in [0,1]$. This is substantially more involved than the expected reward case. 

%Notice that relatively little optimism $\shift = O\left( \ln ( |\Scal||\Acal| / \delta)\right)$ is sufficient to absorb any sampling error in the model and ensure optimistic CVaR estimates for all risk levels $\alpha$.

%Most importantly, there is no dependency in $\shift$ on the risk level or the range of the return which indicates that this parameter is likely easy to tune empirically. Note though that the optimistic distribution computed will depend on the range of the return, and we expect the amount of data required to scale with $\alpha$  
% **EB: why is it easy to tune? Do we back that up? There is dependence in the bonus terms...

These results are a key step towards finite-sample analyses. In future work it would be very interesting to obtain a convergence analysis for distributional Bellman optimality operators in general, though this is outside the scope of this current paper. Such a result could lead to sample-complexity guarantees when combined with our existing analysis.

 \section{Algorithm}
 In the policy evaluation case where we would like to compute optimistic estimates of the CVaR of a given observed policy $\pi$, our algorithm essentially performs an approximate version of the optimistic Bellman update $O_\shift \Tcal^{\pi}$ where $\Tcal^{\pi}$ is the distributional Bellman operator. % from Section~\ref{sec:background}.
%This operator can be composed by distributional Bellman operator $\Tcal^\pi$ which results in optimistic Bellman evaluation update . We further analyze the theoretical properties of this operator and show that an optimistic estimate of $\cvar$ can be obtained with empirical transition kernel $\hat P$ and reward distribution $F_{\hat R}$. 
% Par 2: Control setting:
For the control case where we would like to learn a policy that maximizes CVaR, we instead define a distributional Bellman optimality operator $\Tcal_{\alpha}$.  Analogous to prior work \cite{bellemare2017distributional}, $\Tcal_{\alpha}$ is any operator  that satisfies
 $\Tcal_{\alpha} Z = \Tcal^\pi Z$ for some policy $\pi$ that is greedy w.r.t. CVaR at level $\alpha$. 
 Our algorithm then performs an approximate version of the optimistic Bellman backup  $O_\shift \Tcal_\alpha$, shown in Algorithm~\ref{alg:mdpalg}.

% C51
The main structure of our algorithm resembles categorical distributional reinforcement learning (C51) ~\cite{bellemare2017distributional}. In a similar vein, our algorithm also maintains a return distribution estimate for each state-action pair, represented as a set of $N$ weights $p_i(s,a)$ for $i \in [N]$. These weights represent a discrete distribution with outcomes at $N$ equally spaced locations $z_0 < z_1 < \dots < z_{N-1}$, each $\Delta z = \frac{\Vmax - \Vmin}{N-1}$ apart. The current probability assigned to outcome $z_i$ in $(s,a)$ is denoted by $p_i(s, a)$, where the atom probabilities $p_{1:N}(s,a)$ are given by a differentiable model such as a neural network, similar to C51.  Note that other parameterized representations of the weights \cite{bellemare2017distributional} are straightforward to incorporate.

The main differences between Algorithm~\ref{alg:mdpalg} and existing distributional RL algorithms (e.g. C51) are highlighted in red. We first apply an optimism operator to our successor distribution $F_{Z(s_{t+1}, a)}$ (Lines~\ref{lin:empcdf}--\ref{lin:optcdf}) to form an optimistic CDF $\tilde F_{Z(s_{t+1}, a)}$ for all actions $a\in\Acal$. This operator should encourage exploring actions that might lead to higher CVaR policies for our input $\alpha$. These optimistic CDFs are also used to decide on the successor action in the control setting (Line~\ref{lin:actionchoice}). Then, similar to C51 we apply the Bellman operator $\hat\Tcal z_i$ for $i \in [N]$ and distribute the probability of $\tilde p_i$ to the immediate neighbours of  $\hat\Tcal z_i$, where we calculate the probability mass $\tilde p_i$ with the optimistic CDF $\tilde F_{Z(s_{t+1}, a^\star)}$ (Line~\ref{lin:pdf_from_cdf}). 

%We also use these optimistic CDFs  to decide on the successor action in the target, and for control decisions: in both cases we pick the action that is greedy with respect to the CVaR of the optimistic CDF (see Line~\ref{lin:actionchoice}).

%All actions executed by the algorithm are also chosen as in Line~\ref{lin:actionchoice} of Algorithm~\ref{alg:mdpalg}. For the current state, the algorithm computes the optimistic CDF from the current distribution estimate for each action and then picks the action that achieves the highest $\cvar$.

\begin{algorithm}[!tb]
\KwIn{Parameters: $\gamma$, risk level $\alpha \in [0, 1]$, $c \geq 0$, density model $\rho$, }
\For{t=1, \dots}{
Observe transition $s_t$, $a_t$, $r_t$, $s_{t+1}$\;
\For{$a' \in \Acal$}{
\tcc{emp. CDF of return for $(s_{t+1}, a')$}
\textcolor{red}{$\hat F^{a'}(x) := \sum_{j=0}^{N-1} p_j(s_{t+1}, a') \one \{ x \geq z_j\}$}\;\label{lin:empcdf} 
\tcc{Pseudo-counts using density model}
$\hat n = \frac{1}{\exp(\kappa t^{-1/2} \alpha(\nabla \log \rho_{\theta}(s_{t+1}, a'))^2) - 1}$
 \tcc{Optimistic CDF}
\textcolor{red}{$\tilde F^{a'}(x) := \left[\hat F^{a'}(x) - \frac{c\one\{x \in [\Vmin,\Vmax)\}}{\sqrt{\hat n}} \right]^{+}$}\;\label{lin:optcdf}
}
\begin{minipage}{0.45\textwidth}
\If{Control}{
    $a^\star \gets \argmax_{a \in \Acal} \textcolor{red}{\cvar_{\alpha}(\tilde F^a)}$\label{lin:actionchoice} %\tcp{Choose CVaR-greedy action}
    }
    \end{minipage}\hfill
    \begin{minipage}{0.45\textwidth}
\If{Evaluation}{
$a^\star \sim \pi(.|s_{t+1})$ %\tcp{Choose CVaR action based on policy $\pi$}
}
\end{minipage}\\

     $m_i=0$ for  $i \in \{0,\dots,N-1\}$
     \label{lin:m_start}
     \;
     \For{$j \in 0, \dots, N-1$}{
     \tcc{optimistic PDF from opt. CDF}
     \textcolor{red}{$\tilde p_j \gets \tilde F^{a^\star}\left(z_j + \frac{\Delta z}{2}\right) - \tilde F^{a^\star}\left( z_j - \frac{\Delta z}{2}\right)$\;} \label{lin:pdf_from_cdf}
     \tcc{Project on support of $\{z_i\}$}
        $\tilde{\Tcal}z_j \gets [r_t + \gamma z_j]_{\Vmin}^{\Vmax}$\;
         \tcc{Distribute prob. of $\tilde \Tcal(z_j)$}
        $b_j \gets (\tilde \Tcal z_j - \Vmin)/(\Delta z)$\;
         $l \gets \left \lfloor{b_j}\right \rfloor;
         \qquad u \gets \left \lceil{b_j}\right \rceil$\;
        
        $m_l \gets m_l + \tilde{p}_j(u-b_j)$\;
        $m_u \gets m_u + \tilde{p}_j(b_j-l)$\;
        \label{lin:m_end}
     }
     \textrm{Update return weights $p_{1:N}$ by optimization step on cross-entropy loss} 
     $- \sum_{j=0}^{N-1} m_j \log p_j(s_t, a_t)$
     \label{lin:gradstep}\;
     \tcc{Take next action}
     $a_{t+1} \gets a^\star$ \;
     Update density model for $\rho$ with additional observation of $(s_{t+1}, a_{t+1})$\;
}
 \caption{CVaR-MDP}
      \label{alg:mdpalg}
\end{algorithm}

%\subsection{Function Approximation}
%Algorithm~\ref{alg:mdpalg} is formulated for tabular MDPs for simplicity. However, we can easily extend it to the function approximation setting by making two changes:

Following \cite{bellemare2017distributional}, we train this model using the cross-entropy loss, which for a particular state transition at time $t$ is
\begin{align}
    - \sum_{j=0}^{N-1} m_j \log p_{j}(s_t, a_t)
\end{align}
where $m_{0:N-1}$ are the weights of the target distribution computed in Lines~\ref{lin:m_start}--\ref{lin:m_end} in Algorithm~\ref{alg:mdpalg}. In the tabular setting we can directly update the probability mass $p_j$ by
\begin{align*}
    p_{j}(s_t,a_t) = (1-\beta) p_{j}(s_t,a_t) + \beta m_j(s_t,a_t)
\end{align*}
where $\beta$ is the learning rate.

In tabular settings, the counts n(s,a) can be directly stored and used; however, this is not the case in continuous settings. For this reason, we adopt the pseudo-count estimation method proposed by \cite{ostrovski2017count} and replace $n(s,a)$ by a pseudo-count $\hat N_t(s,a)$ in the optimistic distributional operator (Equation \ref{eq:opt_op}).  Let $\rho$ be a density model and $\rho_t(s,a)$ the probability assigned to the state action pair $(s,a)$ by the model after $t$ training steps. The prediction gain $PG$ of $\rho$ is defined 
\begin{align}
    PG_t(s,a) = \log \rho'_t(s,a) - \log \rho_t(s,a) 
\end{align}
Where $\rho'_t(s,a)$ is the probability assigned to $(s,a)$ if it were trained on that same $(s,a)$ one more time. Now we define the pseudo count of $(s,a)$ as
\begin{align}
    \hat{N}_t(s,a) = (\exp(\kappa t^{-\frac{1}{2}}(PG(s,a))_+ -1)^{-1}
\end{align}
where $\kappa$ is a constant hyper-parameter, and $(PG(s,a))_+$ thresholds the value of the prediction gain at 0. 

Our setting differs from \cite{ostrovski2017count} in the sense that we have to compute the count before taking the action $a$. A naive way would be to try all actions and train the model to compute the counts but this method is slow and requires the environment to support an undo action. Instead, we can estimate $PG$ for all actions as follows. Consider the density model parametrized by $\theta$, $\rho(s,a;\theta)$. After observing $(s,a)$, the training step to maximize the log likelihood will update the parameters by $\theta' = \theta + \alpha \nabla_{\theta}\log\rho(s,a;\theta)$, where $\alpha$ is the learning rate. So we can approximate the new log probability using a first-order Taylor expansion
\begin{align*}
    \log \rho'_t(s,a) 
    &= \log \rho(s,a;\theta')\\ 
    &\approx \log \rho(s,a;\theta) + \nabla_\theta \log \rho(s,a;\theta) (\theta'-\theta) \nonumber\\
    &= \log \rho(s,a;\theta) + \alpha (\nabla_\theta \log \rho(s,a;\theta))^2. 
\end{align*}
This calculation suggests that the prediction gain can be estimated just by computing the gradient of the log likelihood given a state-action pair, i.e., 
$PG(s,a) \approx \alpha (\nabla_\theta \log \rho(s,a;\theta))^2$. As discussed in \cite{graves2017automated} this estimate of prediction gain is biased, but empirically we have found this method to perform well. 

\section{Experimental Evaluation}

We validate our algorithm empirically in three simulated environments against baseline approaches. 
%against existing baselines. In the following, we first provide a description of the benchmark tasks and why we chose them, then discuss baseline methods and further experimental setup before presenting the quantitative results. 
%\subsection{Simulation Domains}
Finance, health and operations are common areas where risk-sensitive strategies are important, and we focus on two health domains and one operations domain. Details, where omitted, are provided in the supplemental material.
%Each of our benchmarks is a simulated task motivated by high-stakes scenarios where one would care about finding risk-sensitive policies: a machine replacement task, a HIV treatment task and an insulin pump regulation task for type 1 diabetes patients.

\paragraph{Machine Replacement} 
%\chris{why: established benchmark in risk-aware lit, tabular and small, can explicitly compute optimal policies, significant difference between cvar and expected ret optimal policy + Add photo of this MDP}

\begin{figure}[tb]
    \centering
    \includegraphics[width=0.5\linewidth]{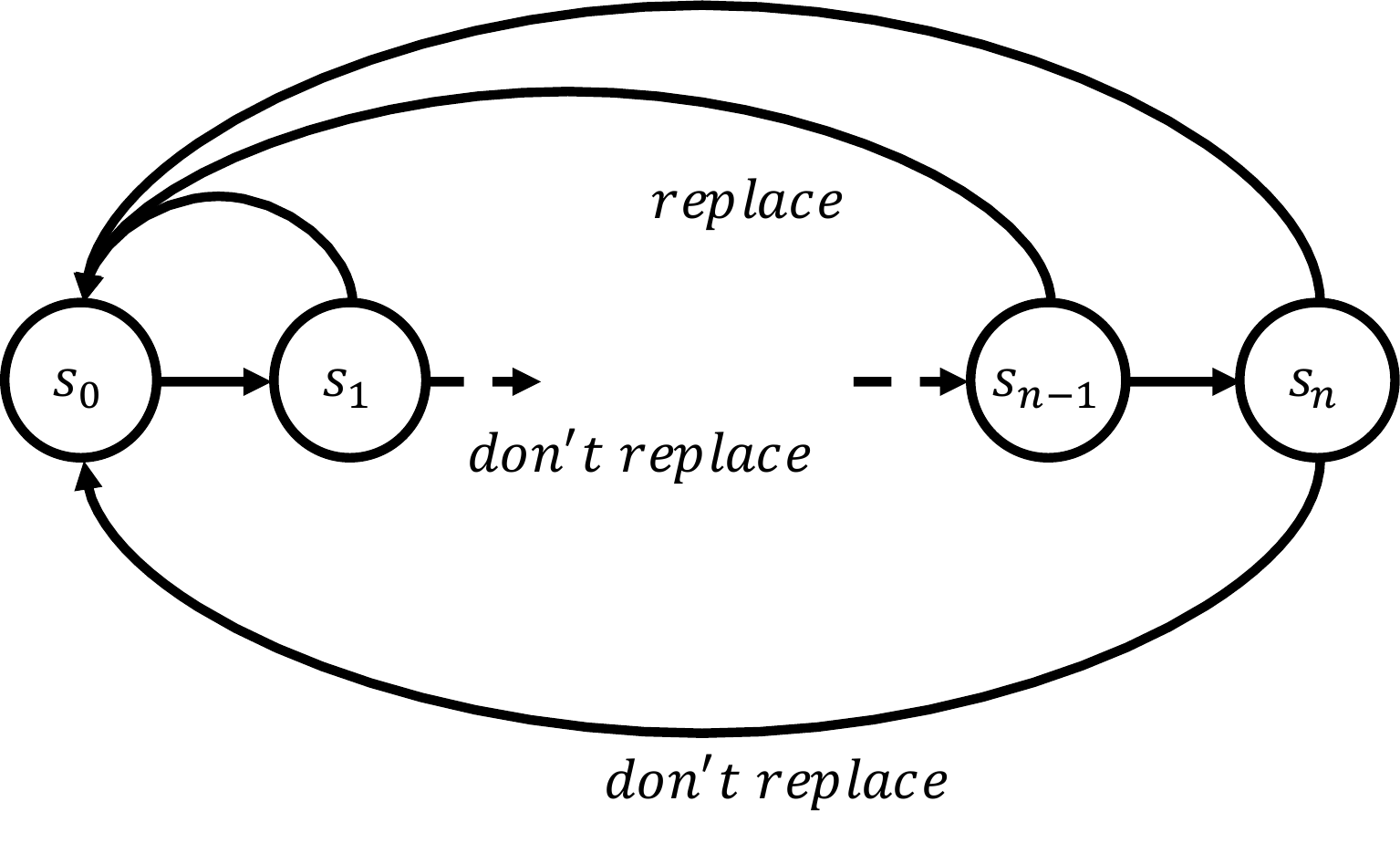}
    \caption{Machine Replacement: This environment consists of a chain of $n$ states, each affording two actions: \emph{replace} and \emph{don't replace}.}
    \label{fig:mdp_mrp}
\end{figure}

Machine repair and replacement is a classic example in the risk sensitive literature, though to our knowledge no prior work has considered how to quickly learn a good risk-sensitive policy for such domains. %% EB: try to add in other refs if possible
Here we consider a minor variant of a prior setting~\cite{delage2010percentile}.
%First we evaluate our algorithm on a variant of the machine replacement environment, an established benchmark in risk sensitive literature ~\cite{delage2010percentile}. 
%Previous work~\cite{delage2010percentile} has focused on risk-sensitive planning for the machine replacement environment, where one tries to find a risk sensitive policy given transition and reward models (with uncertainty). In contrast, here we are concerned with the reinforcement learning setting where the dynamics are unknown. 
Specifically, as shown in Figure \ref{fig:mdp_mrp}, the environment consists of a chain of $n$ (25 in our experiments) states. There are two actions: \emph{replace} and \emph{don't replace} the machine. Choosing \emph{replace} at any state terminates the episode, while choosing \emph{don't replace} moves the agent to the next state in the chain. At the end of the chain, choosing \emph{don't replace} terminates the episode with a high variance cost, and choosing \emph{replace} terminates the episode with a higher cost but lower variance.  This environment is especially a challenging exploration task due to the chain structure of the MDP, as well as the high variance of the reward distributions when taking actions in the last state. Additionally in this MDP it is feasible to exactly compute the $\cvar_{0.25}$-optimal policy, which allows us to compare the learned policy to the true optimal CVaR policy. Note here that the optimal policy for maximizing $\cvar_{0.25}$ is to \emph{replace} on the final state in the chain to avoid the high variance alternative; in contrast, the optimal policy for expected return always chooses \emph{don't replace}.

\paragraph{HIV Treatment}
%\chris{add 1-2 sentences before the description below on why we chose this: mrp is tabular, so we also wanted a continuous state problem, HIV is popular RL task, with limitation that it is deterministic which is not very realistic. To address this, we made stochastic version; this is also a hard exploration problem because we forced the treatment to be the same for 20 days instead of 5}

In order to test our algorithm on a larger continuous state space, we leverage an HIV Treatment simulator. The environment is based on the implementation by \cite{geramifard2015rlpy} of the physical model described in \cite{ernst2006clinical}. The patient state is represented as a $6$-dimensional continuous vector and the reward is a function of number of free HIV viruses, immune response of the body to HIV, and side effects. There are four actions, each determining which drugs are administered for the next $20$ day period: Reverse Transcriptase Inhibitors (RTI), Protease Inhibitors (PI), neither, or both.
%. Typical treatments for HIV patients utilize cocktails consist of one or more RTIs in combination with a PI and patients taking these drugs experience many common and sometimes highly undesirable side effects, often leading to poor compliance. So it is important to minimize the side effects. 
%In this environment there are four actions available (no drug, either one drug, or both) 
There are $50$ time steps in total per episode, for a total of $1000$ days. We chose here a larger number of days per time step compared to the typical setup  ($200$ steps of $5$ days each) to facilitate faster experimentation. This design choice also makes the exploration task harder, since taking one wrong action can drastically destabilize a patient's trajectory. The original proposed model was deterministic, which makes the CVaR policy identical to the policy optimizing the expected value. Such simulators are rarely a perfect proxy for real systems, and in our setting we add Gaussian noise $\sim\mathcal{N}(0, 0.01)$ to the efficacy of each drug (RTI: $\epsilon_1$ and PI: $\epsilon_2$ in \cite{ernst2006clinical}). This change necessitates risk-sensitive policies in this environment.
\begin{figure}[tb]
    \centering
    \includegraphics[width=0.9\linewidth]{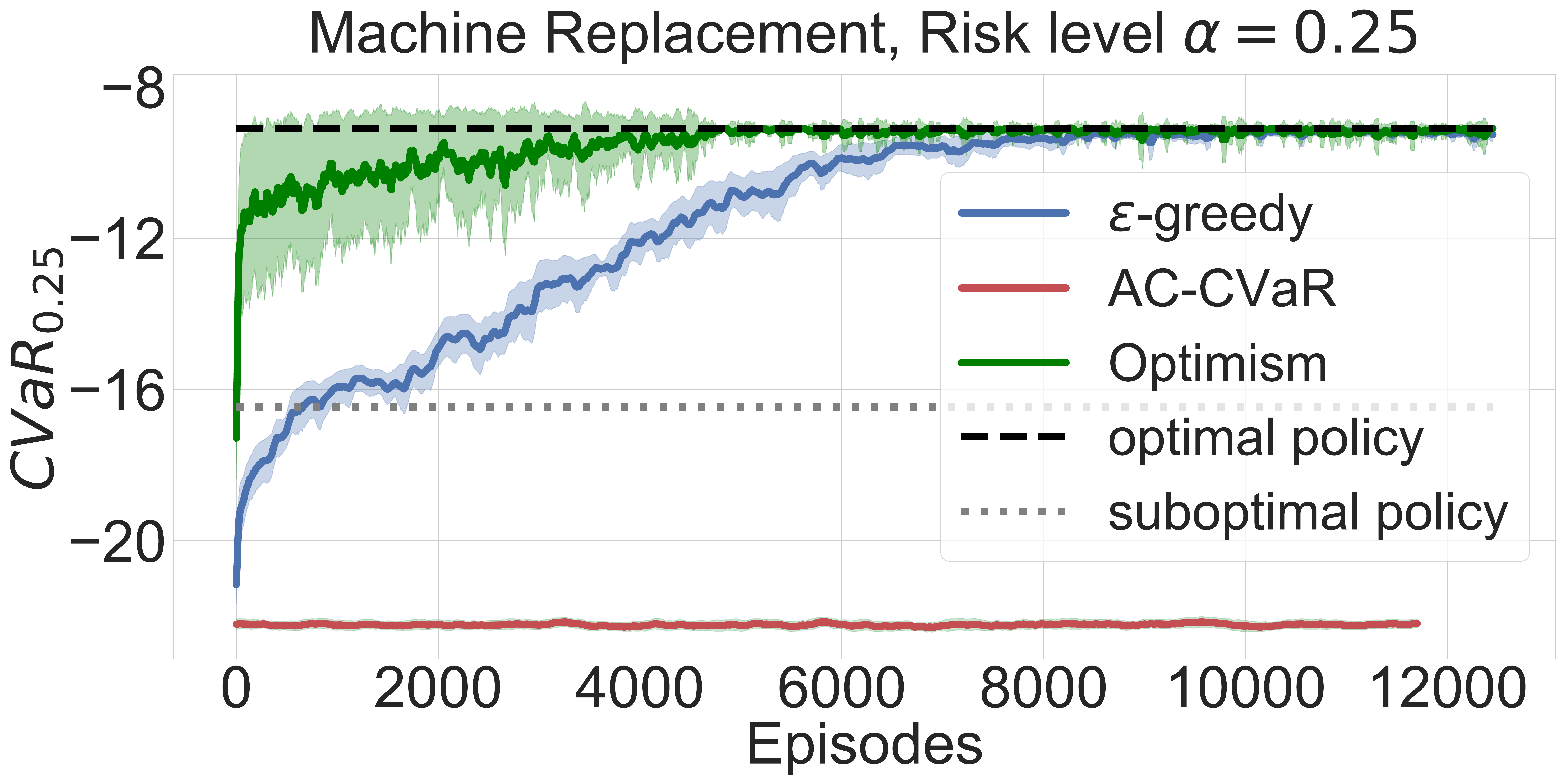}
    \caption{Machine Replacement: The thick grey dashed line is the CVaR$_{0.25}$-optimal policy. The thin dashed lines labeled as the suboptimal policy is the optimal expectation-maximizing policy. The shaded area shows the 95\% confidence intervals.}
    \label{fig:mdp_exp_mrp}
\end{figure}
\begin{figure}[tb]
    \centering
    \includegraphics[width=0.9\linewidth]{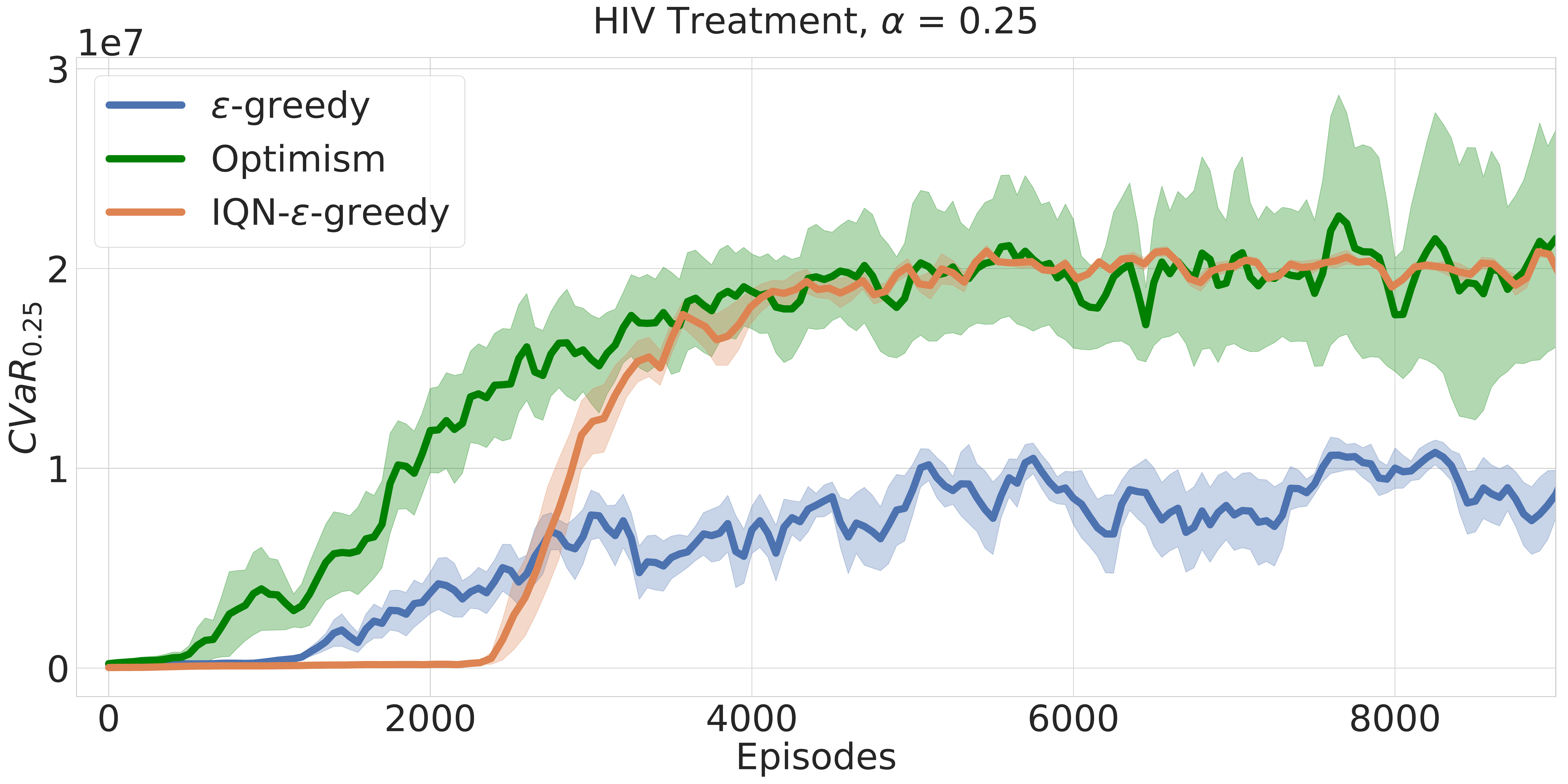}
    \includegraphics[width=0.9\linewidth]{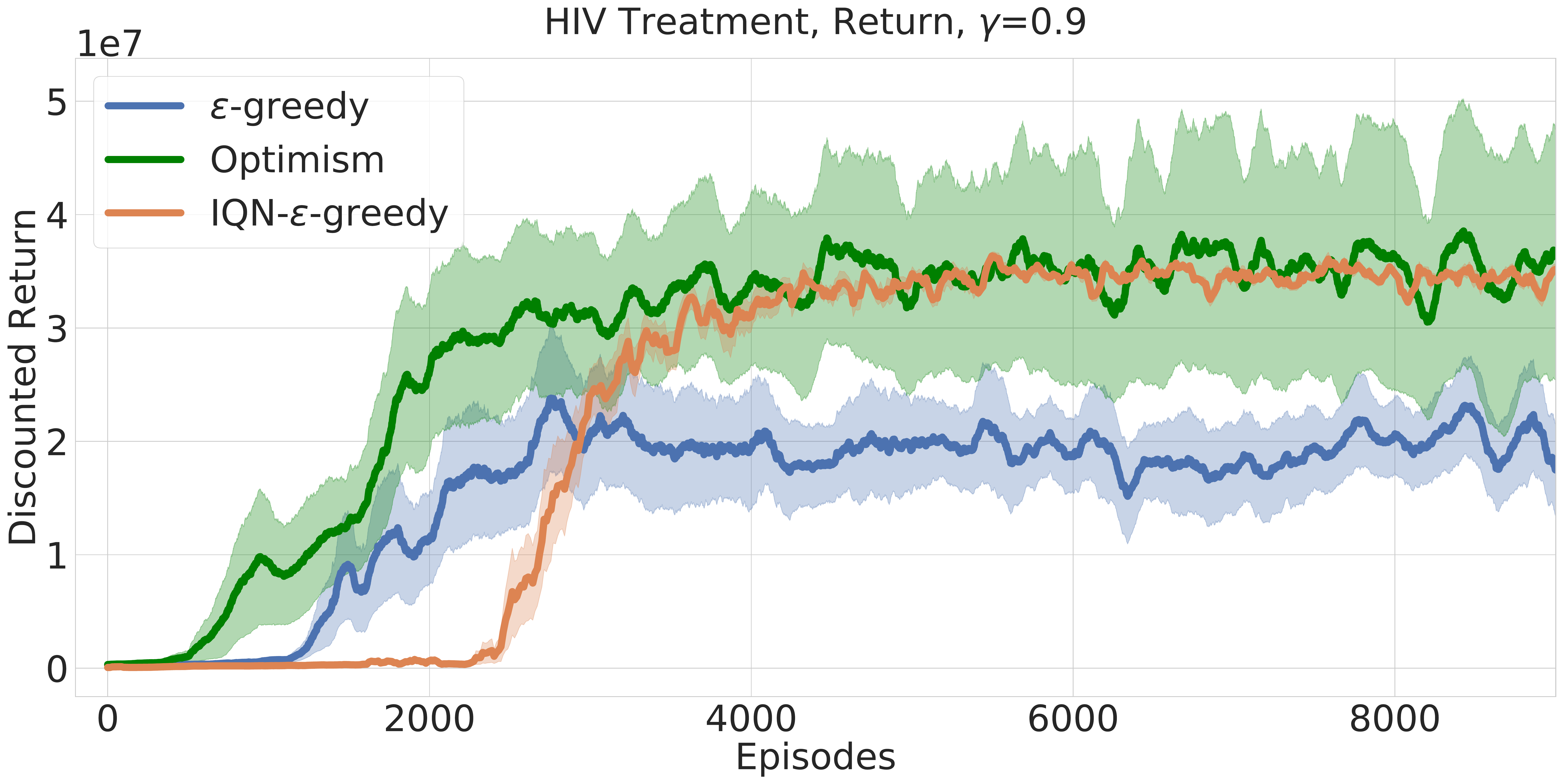}
    \caption{Comparison of our approach against an $\epsilon$-greedy and IQN baseline. All models were trained to optimize the $\cvar_{0.25}$ of the return on a stochastic version of the HIV simulator \cite{ernst2006clinical}. Top: Objective CVaR$_{0.25}$; Bottom: Discounted expected return of the same policies as in top plot.}
    \label{fig:mdp_exp_hiv}
\end{figure}

\begin{figure*}[!tb]\centering
    \includegraphics[width=0.32\linewidth]{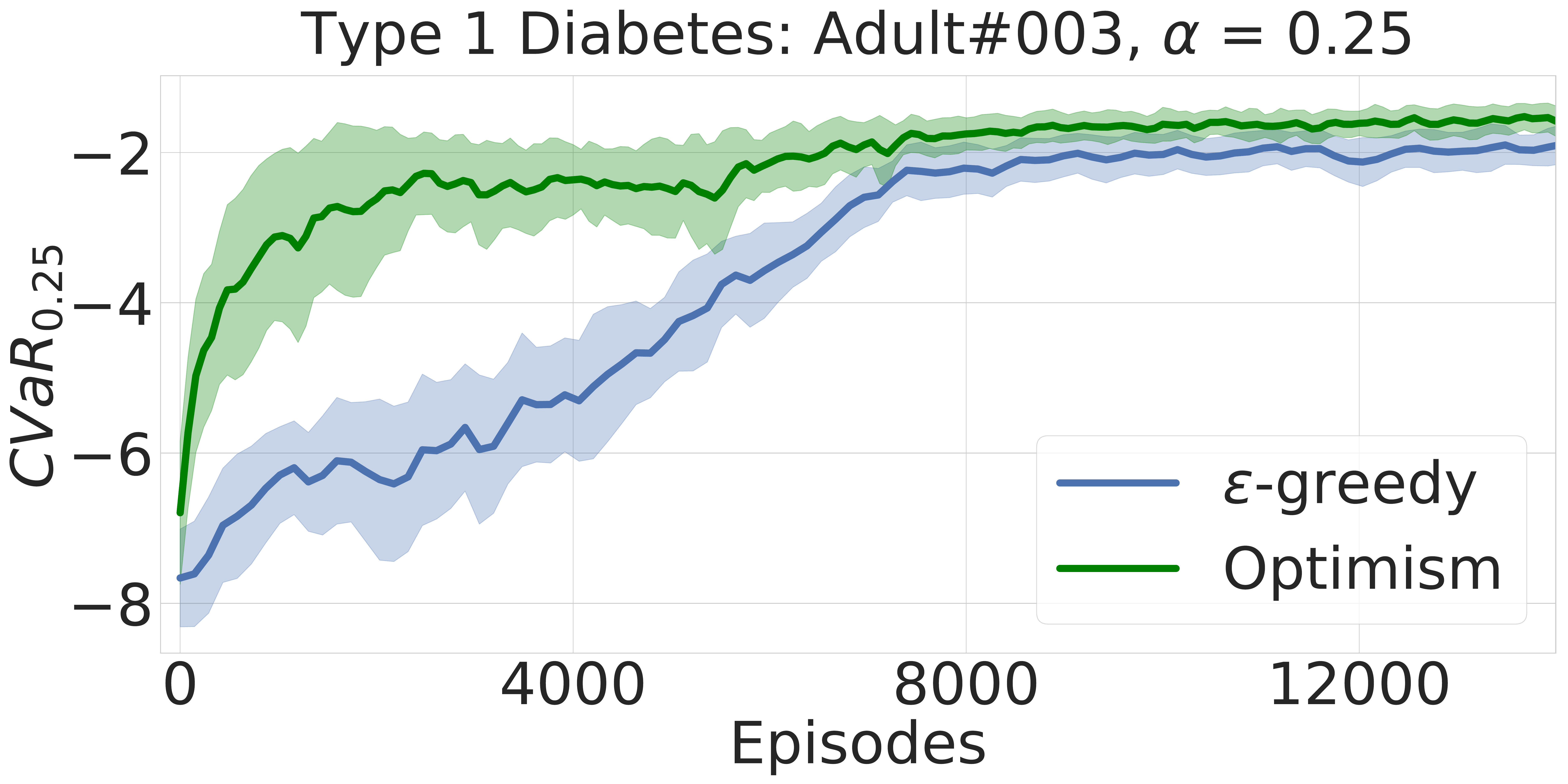}
    \includegraphics[width=0.32\linewidth]{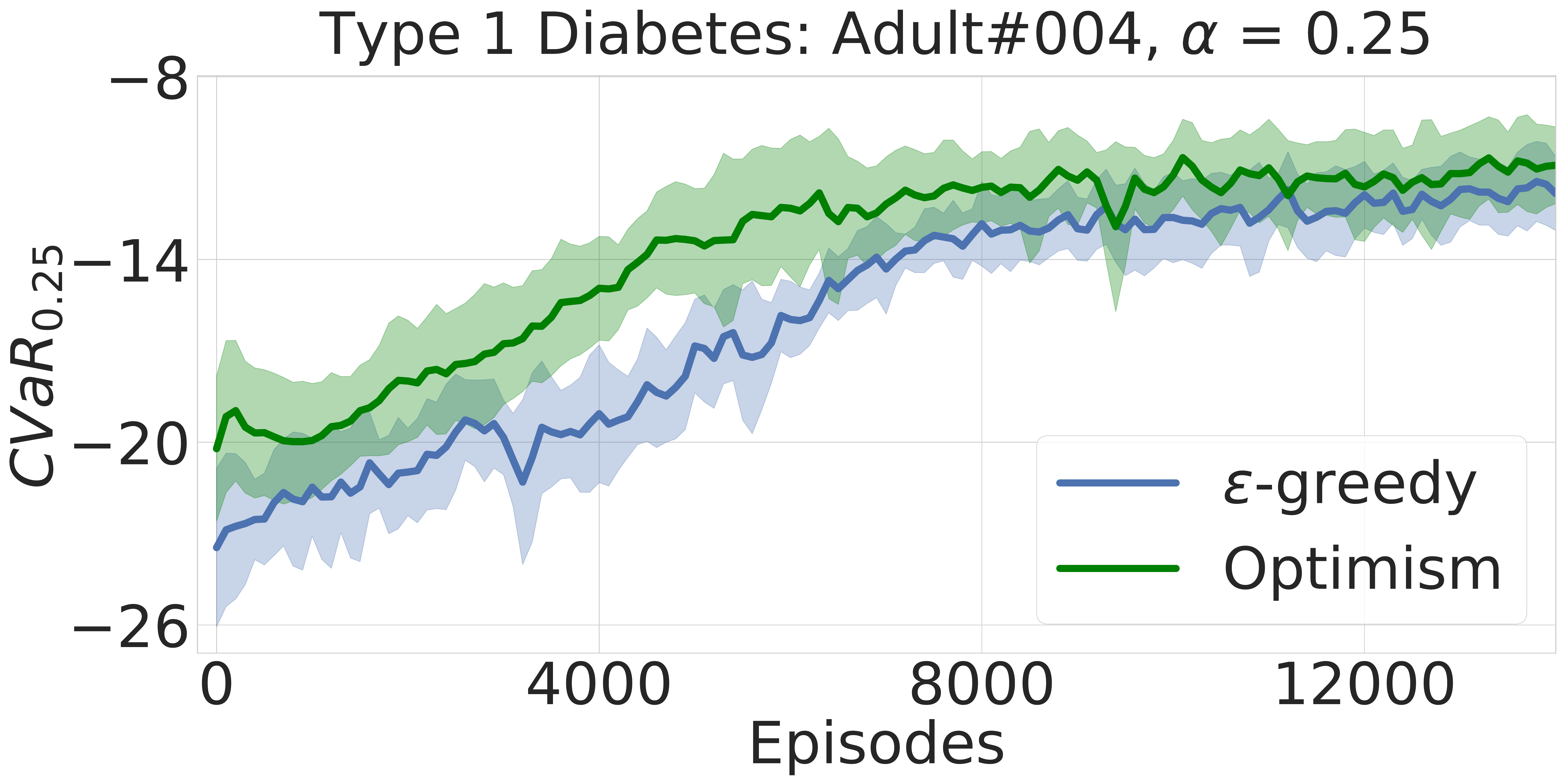}
    \includegraphics[width=0.32\linewidth]{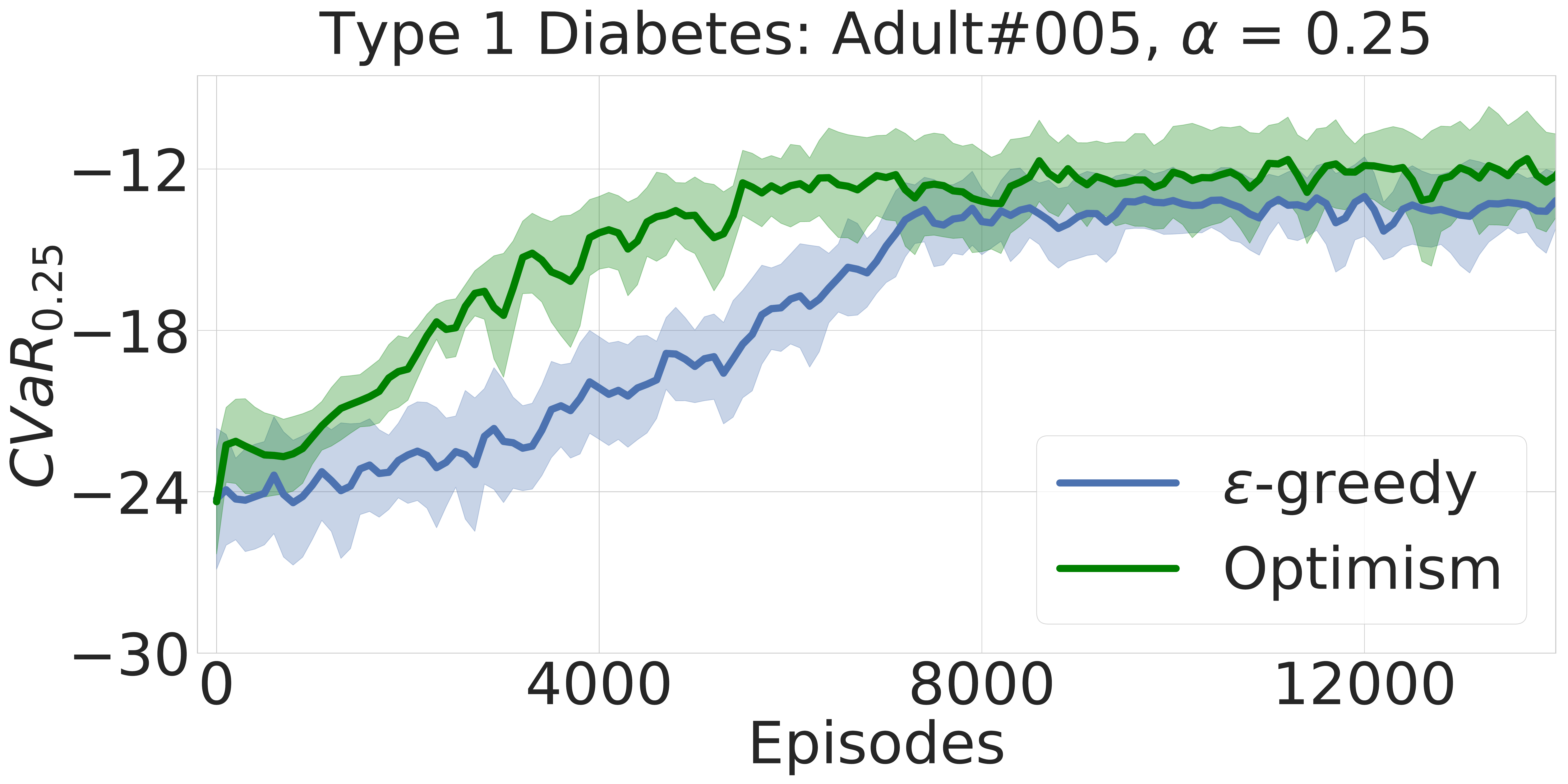}
\caption{Type 1 diabetes simulator: CVaR$_{0.25}$ for three different adults. Plots are averaged over 10 runs with 95\% CI.}
\label{fig:mdp_exp_glucose}
\end{figure*}

%\begin{figure}[!tb]
%\begin{tabular}{l||p{2cm}|p{2cm}|l}
%\toprule
% & $\epsilon$-greedy & $\cvar$-MDP & Reduction  \\ \midrule 
%  Adult\#003 & 11.2\% $\pm$ 3.6\% & \textbf{4.2\% $\pm$ 2.3\%} & 62\%\\ 
%  Adult\#004 & 2.3\% $\pm$ 0.3\% & \textbf{1.4\% $\pm$ 0.6\%}  & 41\% \\ 
%  Adult\#005 & 3.3\% $\pm$ 0.3\% & \textbf{1.7\% $\pm$ 0.6\%} & 47\%\\ 
 %\bottomrule
%\end{tabular}
%\vspace{0.1in}
%\caption{ Type 1 Diabetes simulator, percent of episodes where patients experienced a severe medical condition (hypoglycemia or hyperglycemia), averaged across 10 runs}\label{exp:table:mdp}
%\vspace{-0.2in}
%\end{figure}
 
\begin{figure}[!tb]
\centering
\begin{tabular}{l|p{2.5cm}|p{2.5cm}}
\toprule
     & $\epsilon$-greedy & $\cvar$-MDP   \\ \midrule \midrule 
  Adult\#003 & 11.2\% $\pm$ 3.6\% & \textbf{4.2\% $\pm$ 2.3\%} \\ 
  Adult\#004 & 2.3\% $\pm$ 0.3\% & \textbf{1.4\% $\pm$ 0.6\%}  \\ 
  Adult\#005 & 3.3\% $\pm$ 0.3\% & \textbf{1.7\% $\pm$ 0.6\%} \\ 
 \bottomrule
\end{tabular}
\vspace{0.1in}
\caption{ Type 1 Diabetes simulator, percent of episodes where patients experienced a severe medical condition (hypoglycemia or hyperglycemia), averaged across 10 runs}\label{exp:table:mdp}
\vspace{-0.2in}
\end{figure}

\paragraph{Diabetes 1 Treatment}
%\chris{why: another continuous state task, tuning hyperparameters on same task can be bad(cite Joelle's guidelines), so we introduce a new task\\ what: add to the existing description: size of the state and action space, episode length, what is the noise level? not as hard of an exploration problem}
 
Patients with type 1 diabetes regulate their blood glucose level with insulin in order to avoid  hypoglycemia or hyperglycemia (very low or very high blood glucose level, respectively). 
%We introduce a new benchmark for risk-sensitive RL based on dispensing insulin in a type 1 diabetes simulator \cite{man2014uva}. 
A simulator has been created~\cite{man2014uva} that is an open source version of a simulator that was approved by the FDA as a substitute for certain pre-clinical trials. The state is continuous-valued vector of the current blood glucose level and the amount of carbohydrate intake (through food). The action space is discretized into 6 levels of a bolus insulin injection. The reward function is defined similar to the prior work \cite{bastani2014model} as following:
\begin{equation*}
    r(bg)=
    \begin{cases}
    -\frac{(bg' - 6)^2}{5} & if\ bg'<6 \\
    
    -\frac{(bg' - 6)^2}{10} & if\ bg' \geq 6 
  \end{cases}
\end{equation*}
Where $bg' = bg/18.018018$ which is the estimate of bg (blood glucose) in mmol/L.

Additionally we inject two source of stochasticity into the taken action: First, we add Gaussian noise $\Ncal(0, 1)$ to the action. Second, we delay the time of the injection by at most 5 steps, where the probability of injection at time $t$ is higher than time $t+i, i \geq 1$ following the power law. Each simulation lasts for 200 steps, during which a patient eats five meals. The agent chooses an action after each meal, and after the 200 steps each patient resets to its initial state. 

This domain also readily offers a suite of related tasks, since the environment simulates 30 patients with slightly different dynamics. Tuning hyper-parameters on the same task can be misleading \cite{henderson2018deep}, as is the case in our two previous benchmarks. In this setting we tune baselines and our method on one patient, and test the performance on different patients.

\subsection{Baselines and Experimental Setup}
%\chris{hyperparameter tuning, 1 sentence about architecture + density model, tuning in diabetes case, CVaR-AC, discuss implementation of it;} \rk{ Chris: What does it mean? briefly discuss of why evaluating expected return methods is not useful}

The majority of prior risk-sensitive RL work has not focused on efficient exploration, and there has been very little deep distributional RL work focused on risk sensitivity. Our key contribution is to evaluate the impact of more strategic exploration on the efficiency with which a risk-sensitive policy can be learned. We compare to following approaches:
\begin{enumerate}
    \item $\epsilon$-greedy CVaR: In this benchmark we use the same algorithm, except we do not introduce an optimism operator, instead using an $\epsilon$-greedy approach for exploration. This benchmark can be viewed as analogous to the distributional RL methods of C51~\cite{bellemare2017distributional} if the computed policy had optimized for CVaR instead of expected reward.
    \item IQN-$\epsilon$-greedy CVaR: In this benchmark we use implicit quantile network (IQN) that also uses $\epsilon$-greedy method for exploration~\cite{dabney2018implicit}. We adopted the dopamine implementation of IQN \cite{castro18dopamine}.
    \item CVaR-AC: An actor-critic method proposed by \cite{chow2014algorithms} that maximizes the expected return while satisfying an inequality constraint on the $\cvar$. This method relies on the stochasticity of the policy for exploration.
\end{enumerate}

%We compared our algorithm against two different baselines which either optimize for the $\cvar$-optimal policy or constraint the $\cvar$ of the policy. It is important to notice that a 
Note that a comparison to an expectation maximizing algorithm is uninformative since such approaches are maximizing different (non-risk-sensitive) objectives.

%, since their objective is different. First, we used our algorithm with $\epsilon$-greedy exploration to effectively disentangle the effect of our proposed exploration method given the same underlying learning algorithm. Second, we compared against 

All of these algorithms use hyperparameters, and it is well recognized that $\epsilon$-greedy algorithms can often perform quite well if their hyperparameters are well-tuned. To provide a fair comparison, we evaluated across a number of schedules for reducing the $\epsilon$ parameter for both $\epsilon$-greedy and IQN, and a small set of parameters (4-7) for the optimism value $c$ for our method. We used the specification described in Appendix C of \cite{chow2014algorithms} for CVaR-AC.
%, and tune our method with 4 different optimism value $c$.

The system architectures used in continuous settings are identical for Baseline 1 ($\epsilon$-greedy) and our method. This consists of  %the categorical distributional RL or C51 algorithm 
 2 hidden layers of size 32 with ReLU activation for Diabetes 1 Treatment, and 4 hidden layers of size 128 with ReLU activation for HIV Treatment, both followed by a softmax layer for each action. Similarly for IQN we used the same architecture, followed by a cosine embedding function and a fully connected layer of size 128 for HIV Treatment (32 for Diabetes 1 Treatment) with ReLU activation, followed by a softmax layer. The density model is a realNVP \cite{dinh2016density} with 3 hidden layers each of size 64.

%For both HIV Treatment and Diabetes 1 Treatment we tune $\epsilon$-greedy with 10 different schedules and our method with 7 different optimism value. Additionally for 

All results are averaged over 10 runs and we report 95\% confidence intervals. We report the performance of $\epsilon$-greedy at evaluation time (setting $\epsilon$ = 0), which is the best performance of $\epsilon$-greedy.

For the Diabetes Treatment domain, hyperparameters are optimized only on \texttt{adult\#001}. We then report results of the methods using those hyperparameters on \texttt{adult\#003} , \texttt{adult\#004} and \texttt{adult\#005}. 

\subsection{Results and Discussion}
%\chris{refer to all plots; consistent improvement, in easy and hard exploration tasks, couldn't get AC method to work, even on the easiest problem, so do not show it on the rest; discuss table for glucose} 

Results on machine replacement environment (Figure \ref{fig:mdp_exp_mrp}), HIV Treatment (Figure \ref{fig:mdp_exp_hiv}) and Diabetes 1 Treatment (Figure \ref{fig:mdp_exp_glucose}) all show 
our optimistic algorithm achieves better performance much faster than the baselines.  

In Machine Replacement (Figure~\ref{fig:mdp_exp_mrp}) we see that our method quickly converges to the optimal CVaR performance. 
%consistent improvement of our method over the baselines. 
Unfortunately despite our best efforts, our implementation of CVaR-AC did not perform well even on the simplest environment, so we did not show the performance of this method on other environments. One challenge here is that CVaR-AC has a significant number of  hyper-parameters, including 3 different learning rates schedule for the optimization process, initial Lagrange multipliers and the kernel functions.

In the HIV Treatment we also see a clear and substantial benefit to our optimistic approach over the baseline $\epsilon$-greedy approach  and IQN(Figure ~\ref{fig:mdp_exp_hiv}).

Figure~\ref{fig:mdp_exp_glucose} is particularly encouraging, as it shows the results for the diabetes simulator across 3 patients, where the hyperparameters were fixed after optimizing for a separate patient. Since in real settings it would be commonly necessary to fix the hyperparameters in advance, this result provides a nice demonstration that the optimistic approach can consistently equal or significantly improve over an $\epsilon$-greedy policy in related settings, similar to the well known results in Atari in which hyperparameters are optimized for one game and then used for multiple others.

%Additionally, in a harder exploration task (Figure \ref{fig:mdp_exp_hiv}: HIV Treatment) the baseline with $\epsilon$-greedy exploration did not match the performance of optimism based exploration and converged to a sub-optimal policy. 

\paragraph{"Safer" Exploration.} Our primary contribution is a new algorithm to learn risk-sensitive policies quickly, with less data. However, an interesting side benefit of such a method might be that the number of extremely poor outcomes experienced over time may also be reduced, not due to explicitly prioritizing a form of safe exploration, but because our algorithm may enable a faster convergence to a safe policy.
%Our method is mainly focused on finding a safe policy fast, an interesting question would be how many catastrophic error our method will make before learning a safe policy. 
To evaluate this, we consider a risk measure proposed by \cite{clarke2009statistical}, which quantifies the risk of a severe medical condition based on how close their glucose level is to hypoglycemia (blood glucose, $\leq$3.9 mmol/l) and hyperglycemia (blood glucose, $\geq$10 mmol/l). %** EB: state what it is

Table \ref{exp:table:mdp} shows the fraction of episodes in which each patient experienced a severely poor outcome for each algorithm while learning. Optimism-based exploration approximately halves the number of episodes with severely poor outcomes, highlighting a side benefit of our optimistic approach of more quickly learning a good safe policy.

%in on average a 62\%, 41\% and 47\% reduction in severe conditions for three different adults while learning a safe policy. 

\section{Related Work}
\label{sec:related}

Optimizing policies for risk sensitivity in MDPs has been long studied, with policy gradient \cite{sampling,tamar2015policy}, actor critic  \cite{tamar2013variance} and TD methods~\cite{tamar2013variance,sato2001td}. While most of this work considers mean-variance trade objectives, \cite{chow2015risk} establish a connection between a optimizing CVaR and robustness to modeling errors, presenting a value iteration algorithm.
%for finding the CVaR-optimal policy. 
In contrast, we do not assume access to transition and rewards models.  \cite{chow2014algorithms} present a policy gradient and actor-critic algorithm for an expectation-maximizing objective with a CVaR constraint. None of these works considers systematic exploration but rely on heuristics such as $\epsilon$-greedy or on the stochasticity of the policy for exploration. Instead, we focus on how to explore systematically to find a good CVaR-policy.

Our work builds upon recent advances on distributional RL~\cite{bellemare2017distributional,rowland2018analysis,dabney2018distributional} which are still concerned with optimizing expected return. Notably, \cite{dabney2018implicit} aims to train risk-averse and risk-seeking agents, but does not address the exploration problem or attempts to find optimal policies quickly.

\cite{dilokthanakul2018deep} uses risk-averse objectives to guide exploration for good performance w.r.t. expected return.
\cite{moerland2018potential} leverages the return distribution learned in distributional RL as a means for optimism in deterministic environments. \cite{mavrin2019distributional} follow a similar pattern but can handle stochastic environments by disentangling intrinsic and parametric uncertainty. While they also evaluate the policy that picks the VaR-greedy action in one experiment, their algorithm still optimizes expected return during learning.  In general, these approaches are fundamentally different from ours which learns CVaR policies in stochastic environments efficiently by introducing optimism \emph{into} the learned return distribution.

\section{Conclusion}
\label{conclusion}
We present a new algorithm for quickly learning CVaR-optimal policies in Markov decision processes. This algorithm is the first to leverage optimism in combination with distributional reinforcement learning to learn risk-averse policies in a sample-efficient manner. Unlike existing work on expected return criteria which rely on reward bonuses for optimism, We introduce optimism by directly modifying the target return distribution and provide a theoretical justification that in the evaluation case for finite MDPs, this indeed yields optimistic estimates. We further empirically observe significantly faster learning of CVaR-optimal policies by our algorithm compared to existing baselines on several benchmark tasks. This includes simulated healthcare tasks where risk-averse policies are of particular interest: HIV medication treatment  and insulin pump control for diabetes type 1 patients. 

\section{Acknowledgments}
The research reported here was supported by NSF CAREER award, ONR Young Investigator award and Microsoft faculty fellowship.
\bibliography{refs.bib}
\bibliographystyle{aaai}
\newpage
\onecolumn
\appendix
\section{Appendix}
\subsection{Experimental Details}
\subsubsection{Machine Replacement}
\label{appx:mrp}
Machine replacement environment consist of $n$ states (we use $n=25$ in the experiment), where action \emph{replace} transitions to the terminal state with cost $\mathcal{N}(\mu_{r,t}, \sigma_{r,t})$ at state $t$, where $\mu_{r,t} = r_{max} - \frac{t}{n}(r_{max}-r_{min})$ and $\sigma_{r,t} = 0.1 + 0.01 t$. Action \emph{don't replace} has cost $\mathcal{N}(0, 1e-2)$ and transitions to state $t+1$. In our experiment we used $r_{max}=23, r_{min}=10$. However, for the last state $n$ action \emph{don't replace} has cost $\mathcal{N}(\mu_{r}, 10)$, where we used $\mu_r = 8$, and transitions to the terminal state. For C51 algorithm we use $V_{min}=-50, V_{max}=50$, $\gamma=0.99$, learning rate $0.01$ and 51 atoms.

\textbf{Tuning:}  We use $\epsilon$-greedy with schedule $(\text{start}, \text{end}, \text{n}) = (0.9, 0.1, 5000)$ that starts with $\epsilon = \text{start}$ and decays linearly to $\epsilon = \text{end}$ in $n$ time steps, staying constant afterwards. This schedule achieved the best performance in our experiments when compared to other linear schedules \{$(0.9, 0.3, 5000)$, $(0.9, 0.1, 10000)$, $(0.9, 0.1, 15000)$, $(0.9, 0.05, 5000)$\}, and exponential decays with schedule in the form of ($\epsilon_0$, $d$, $step$): \{(0.9, 0.99, 5), (0.9, 0.99, 20), (0.9, 0.99, 2), (0.9, 0.99, 30), (0.5, 0.99, 5)\} where $\epsilon = \epsilon_0 \times d^{episode/step}$. We have also tried our algorithm with optimism values of $c = [0.25, 0.5, 1, 2]$.

For the actor critic method we use the CVaR limit as -10, radial basis function as kernel and other set of hyper-parameters are as described in the appendix of \cite{chow2014algorithms}.

\begin{figure}[b]
\minipage{0.48\textwidth}
  \includegraphics[width=\linewidth]{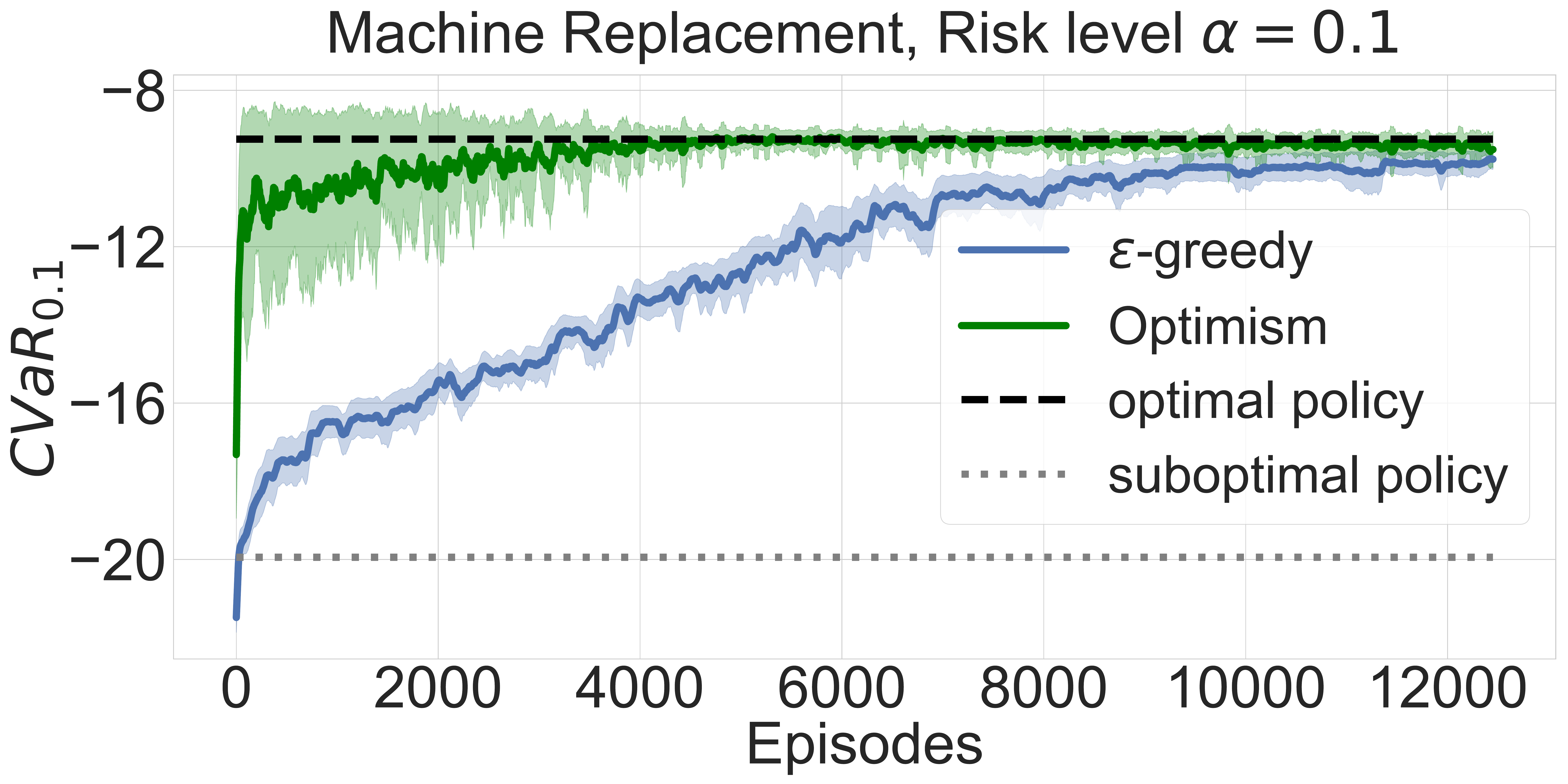}
\endminipage\hfill
\minipage{0.48\textwidth}
  \includegraphics[width=\linewidth]{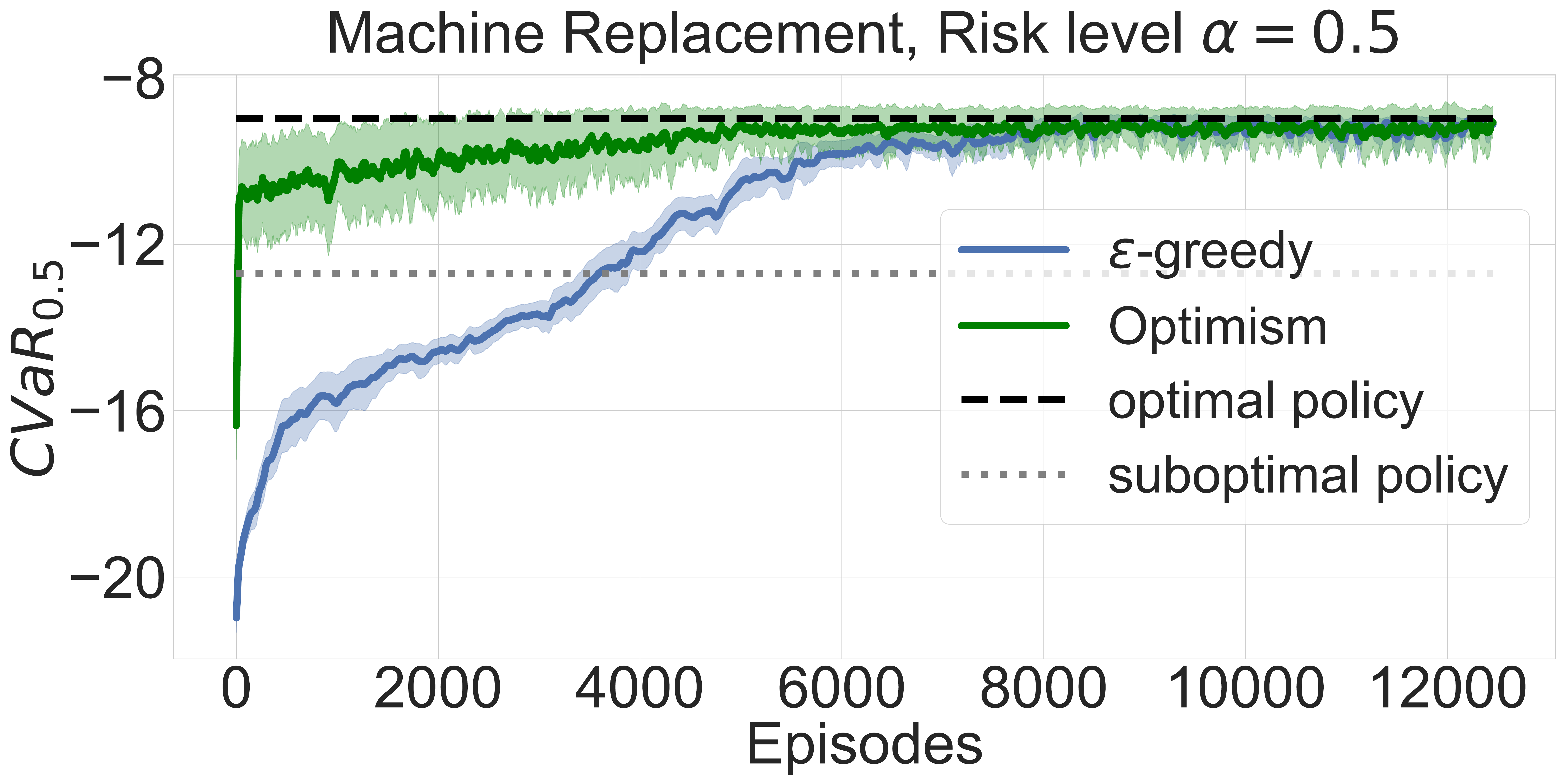}
\endminipage\hfill
\caption{Machine Replacement with different risk levels. Left: risk level $\alpha=0.1$, Right: risk level $\alpha=0.5$}
\label{fig:appx:mrp}
\end{figure}

\textbf{Additional Experiments:} Additional to the risk level $\alpha=0.25$, we observe the same gain in the performance for other risk levels. As shown in figure \ref{fig:appx:mrp}, optimism based exploration shows a significant gain over $\epsilon-$greedy exploration for risk levels $\alpha=0.1$ and $\alpha=0.5$. 

\subsubsection{Type 1 Diabetes Simulator}\label{appx:simglucose}
An open source implementation of type 1 diabetes simulator \cite{simglucose} simulates 30 different virtual patients, 10 child, 10 adolescent and 10 adult. For our experiments in this paper we have used \texttt{adult\#003}, \texttt{adult\#004} and \texttt{adult\#005}. Additionally we have used \texttt{"Dexcom"} sensor for CGM (to measure blood glucose level) and \texttt{"Insulet"} as a choice of insulin pump. All simulations are 10 hours for each patient and after 10 hour, patient resets to the initial state. Each step of simulation is 3 minutes.

State space is a continous vector of size 2 (glucose level, meal size) where glucose level is the amount of glucose measured by \texttt{"Dexcom"} sensor and meal size is the amount of Carbohydrate in each meal. 

Action space is defined as (bolus, basal=0) where amount of bolus injection discretized by 6 bins between 30 (max bolus, a property of the \texttt{"Insulet"} insulin pump) and 0 (no injection). Additionally we inject two source of stochasticity to the taken action, assume action $a=(a_b, 0)$ at time $t$ is the agent's decision, then we take the action $a=(a'_b,0)$ at time $t'$ where:
\begin{align*}
    a'_b &= a_b + \Ncal(0, 1)\\
    t' &= t + c - \left \lfloor{x \times c}\right \rfloor 
\end{align*}
Where $x \sim P(x;1) = 2 x^{-1}$ is drawn from the power law distribution and $c = 5$. Note that this means delay the action at most 5 step where the probability of taking the action at time $t$ is higher than time $t+i, i \geq 1$ following the power law. Since each step of simulation is 3 minutes, patient might take the insulin up to 15 minutes after the prescribed time by the agent. 

Reward structure is defined similar to the prior work \cite{bastani2014model} as following:
\begin{equation*}
    r(bg)=
    \begin{cases}
    -\frac{(bg' - 6)^2}{5} & if\ bg'<6 \\
    
    -\frac{(bg' - 6)^2}{10} & if\ bg' \geq 6 
  \end{cases}
\end{equation*}
Where $bg' = bg/18.018018$ which is the estimate of bg (blood glucose) in mmol/L rather than mg/dL. Additionally if the amount of glucose is less than 39 mg/dL agent incurs a penalty of $-10$. 

We generated a meal plan scenario for all the patients that is meal of size 60, 20, 60, 20 CHO with the schedule 1 hour, 3 hours, 5 hours and 7 hours after starting the simulation. Notice that this will make the simulation horizon 200 steps and 5 actionable steps (initial state, and after each meal).

\textbf{Categorical Distributional RL:} The C51 model consist of 2 hidden layers each of size 32 and ReLU activation function, followed by  of $|\Acal|$ each with 51 neurons followed by a softmax activation, for representing the distribution of each action. 

We used Adam optimizer with learning rate $1e-3$, $\beta_1=0.9$,$\beta_2=0.999$ and $\epsilon=1e-8$. We set $V_{max}=15$, $V_{min}=-40$, 51 probability atoms, and used batch size of 32. For computing the CVaR we use 50 samples of the return.

\textbf{Density Model:} For log likelihood density model we used realNVP \cite{dinh2016density} with 3 layers each of size 64. The input of the model is a concatenated vector of $(s,a)$. We used same hyper parameters for optimizer as in C51 model. We have used constant $\kappa = 1e-5$ for computing the pseudo-count.

\textbf{Tuning:} We have tuned our method and $\epsilon$-greedy on patient \texttt{adult\#001} and used the same parameters for the other patients. We tried 5 different linear schedule of $\epsilon$-greedy, \{(0.9, 0.1, 2), (0.9, 0.05, 4), (0.9, 0.05, 6), (0.9, 0.3, 4), (0.9, 0.3, 4),  (0.9, 0.05, 10)\} where first element is the initial $\epsilon$, second element is the final $\epsilon$ and the third element is episode ratio (i.e. epsilon starts from initial and reaches to the final value in episode ratio fraction of total number of episodes, linearly). Additionally we have tried 5 different exponential decay schedule for $\epsilon$-greedy in the form of ($\epsilon_0$, $d$, $step$):  \{(0.9, 0.99, 5), (0.9, 0.99, 20), (0.9, 0.99, 2), (0.9, 0.99, 30), (0.5, 0.99, 5)\} where $\epsilon = \epsilon_0 \times d^{episode/step}$. The first of the exponential decay set preformed the best. We have also tested our algorithm with constant optimism values of $[0.2, 0.4, 0.5, 0.8, 1, 2, 5]$ where we picked the best value $0.5$.

\subsubsection{HIV Simulator}\label{appx:hivexp}
The environment is an implementation of the physical model described in \cite{ernst2006clinical}. The state space is of dimension 6 with and action space is of size 4, indicating the efficacy of being on the treatment. $\epsilon_1, \epsilon_2$ described in \cite{ernst2006clinical} takes values as $\epsilon_1\in \{0, 0.7\}$ and $\epsilon_2\in \{0, 0.3\}$. We have also added the stochasticity to the action by a random gaussian noise. So the efficacy of a drug is computed as $\epsilon_i + \Ncal(0, 0.01)$.

The reward structure is defined similar to the prior work \cite{ernst2006clinical}. And we simulate for 1000 time steps, where agent can take action in 50 steps (each 20 simulation step) and actions remains constant in each interval. While trianing we normalize the reward by dividing them by $1e6$.

\textbf{Categorical Distributional RL:} The C51 model consist of 4 hidden layers each of size 128 and ReLU activation function, followed by  of $|\Acal|$ each with 151 neurons followed by a softmax activation, for representing the distribution of each action. 

We used Adam optimizer with learning rate decay schedule from $1e-3$ to $1e-4$ in half a number of episodes, $\beta_1=0.9$, $\beta_2=0.999$ and $\epsilon=1e-8$. We set $V_{max}=40$, $V_{min}=-10$, 151 probability atoms, and used batch size of 32. For computing the CVaR we use 50 samples of the return.

\textbf{Implicit Quantile Network:} IQN model consists of 4 hidden layers with size 128 and ReLU activation. Then an embedding of size 64 computed by ReLU$(\sum_{i=0}^{n-1} \cos (\pi i \tau) w_{ij} + b_j)$. Then we take the element wise multiplication of the embedding and the output of 4 hidden layers, followed by a fully connected layer with size 128 and ReLU activation, and a softmax layer. We used 8 samples for $N$ and $N'$ and 32 quantiles. 

\textbf{Density Model:} For log likelihood density model we used realNVP \cite{dinh2016density} with 3 layers each of size 64. The input of the model is a concatenated vector of $(s,a)$. We used same hyper parameters for optimizer as in C51 model. We have used constant $\kappa = 1e-5$ for computing the pseudo-count.

\textbf{Tuning:} We have tuned our method, $\epsilon$-greedy and IQN. For $\epsilon$-greedy we tried 5 different linear schedule of $\epsilon$-greedy, \{(0.9, 0.05, 10), (0.9, 0.05, 8), (0.9, 0.05, 5), (0.9, 0.05, 4), (0.9, 0.05, 2)\} where first element is the initial $\epsilon$, second element is the final $\epsilon$ and the third element is episode ratio (i.e. epsilon starts from initial and reaches to the final value in episode ratio fraction of total number of episodes, linearly). Additionally we have tried 5 different exponential decay schedule for $\epsilon$-greedy and IQN in the form of ($\epsilon_0$, $d$, $step$):  \{(1.0, 0.9, 10), (1.0, 0.9, 100), (1.0, 0.9, 500), (1.0, 0.99, 10), (1, 0.99, 100)\} where $\epsilon = \epsilon_0 \times d^{episode/step}$. The first of the linear decay set preformed the best. We have also tested our algorithm with constant optimism values of $(0.2, 0.4, 0.5, 0.8, 1, 2, 5)$ where we picked the best value $0.8$.

%\subsubsection{AC-CVaR} We have also implemented the Actor-Critic algorithm for CVaR optimization method described in \cite{chow2014algorithms}. Where the method maximized the reward given a constraint on the CVaR. Unfortunately our implementation did not perform well on the machine repair environment. We suspect the reason is many hyperparameters that are need to be tuned, including 3 different learning rates schedule for the optimization process, initial Lagrange multipliers and the kernel function. We have also tried the specification described in Appendix C of \cite{chow2014algorithms}, and the shown plot in the main text are generated with those parameters. 
\subsection{Theoretical Analysis}

\begin{proposition}[Restatement of Proposition~\ref{mdp:contraction}]
    For any $\shift$, the $O_\shift$ operator is a non-expansion in the Cram\'er distance $\bar\ell_2$. This implies that optimistic distributional Bellman backups $O_\shift \Tcal^\pi$ and the projected version $\Pi_{\Ccal} O_\shift \Tcal^\pi$ are $\sqrt{\gamma}$-contractions in $\bar\ell_2$ and iterates of these operators converge in $\bar\ell_2$ to a unique fixed-point.
\end{proposition}
\begin{proof}
 Consider $Z, Z' \in \Zcal$, any state $s \in \Scal$ and action $a \in \Acal$ with CDFs $F_{Z(s,a)}$ and $F_{Z'(s,a)}$ and consider the application of the optimism operator $O_\shift$: 
    \begin{align}
        \int (F_{O_\shift Z(s,a)}(x) - F_{O_\shift Z'(s,a)}(x))^2 dx
        =     \int_{\Vmin}^{\Vmax} ([F_{Z(s,a)}(x) - \shift]^+ - [F_{Z'(s,a)}(x) - \shift]^+)^2 dx.
    \end{align}
    Generally, for any $a \geq b$ we have
    \begin{align}
        ( [a - \shift]^+ - [b - \shift]^+)^2 = 
        \begin{cases}
            (a - b)^2 & \textrm{if } a,b \geq \shift\\
            (a - \shift)^2 \leq (a - b)^2 & \textrm{if } a > \shift \geq b\\
            0 & \textrm{if } \shift \geq a, b
        \end{cases}
    \end{align}
and applying this case-by-case bound to the quantity in the integral above, we get 
\begin{align}
    \int (F_{O_\shift Z(s,a)}(x) - F_{O_\shift Z'(s,a)}(x))^2 dx
    \leq     \int_{\Vmin}^{\Vmax} (F_{Z(s,a)}(x) - F_{Z'(s,a)}(x))^2 dx.
\end{align}
By taking the square root on both sides as well as a max over states and actions, we get that $O_\shift$ is a non-expansion in $\bar\ell_2$.
The rest of the statement follows from the fact that $\Tcal^\pi$ is a $\sqrt{\gamma}$-contraction and $\Pi_{\Ccal}$ a non-expansion \cite{rowland2018analysis} and the Banach fixed-point theorem.
\end{proof}

\begin{theorem}[Restatement of Theorem~\ref{mdp:optimistic_cvar}]
Let the shift parameter in the optimistic operator be sufficiently large which is $\shift = O\left( \ln ( |\Scal||\Acal| / \delta)\right)$. Then with probability at least $1 - \delta$, the iterates $\cvar_{\alpha}((O_{\shift} \hat \Tcal^\pi)^m Z_0)$ converges for any risk level $\alpha$ and initial $Z_0 \in \Zcal$ to an optimistic estimate of the policy's conditional value at risk. That is, with probability at least $1 - \delta$, 
\begin{align*}
    \forall s,a : \cvar_{\alpha}((O_{\shift} \hat \Tcal^\pi)^\infty Z_0(s,a)) \geq \cvar_{\alpha}(Z_\pi(s,a))
\end{align*}
\end{theorem}
\begin{proof}
     By Lemma~\ref{lem:stochdom_implies_cvaropt} and Lemma~3 by \cite{bellemare2017distributional}, we know that $Z_{i+1} \gets O_\shift\hat \Tcal^\pi Z_i$ converges to a unique fixed-point $Z_\infty$, independent of the initial $Z_0$. Hence, without loss of generality, we can choose $Z_0 = Z_\pi$.
 
      We proceed by first showing how our result will follow under a particular definition of $\shift$, and then show what that definition is. 
      Assume that we have obtained a value for $\shift$ that satisfies the assumption of Lemma~\ref{appx:mdp:opt} (see other parts of this appendix), and let $\tilde Z = Z_i$ and $Z = Z_\pi$. Then Lemma~\ref{appx:mdp:opt} implies that if  $F_{Z_i(s,a)} \leq F_{Z_\pi(s,a)}$ for all $(s,a)$, then also $F_{Z_{i+1}(s,a)} \leq  F_{Z_\pi(s,a)}$ for all $(s,a)$.
     Thus, $F_{Z_\infty(s,a)} \leq F_{Z_\pi(s,a)}$ for all $(s,a)$. Finally, we can use  Lemma~\ref{lem:stochdom_implies_cvaropt} to obtain the desired result of our proof statement,  $\cvar_{\alpha}(F_{Z_\infty(s,a)}) \geq \cvar_{\alpha}(F_{Z_\pi(s,a)})$ for all $(s,a)$.

%    We will apply Lemma~\ref{appx:mdp:opt} with $Z = Z_\pi$ inductively but first w
    
    Going back, we use concentration inequalities to determine the value of $c$ that ensures the required condition in  Lemma~\ref{appx:mdp:opt} (expressed in Eq.~\eqref{eq:assum_opt_bonus}). The DKW-inequality which give us that for any $(s,a)$ with probability at least $ 1 - \delta$
    \begin{align}
         \|F_{R(s,a)} - F_{\hat R(s,a)}\|_\infty & \leq \sqrt{ \frac{1}{2n(s,a)} \ln \frac{2}{\delta}}.\label{eqn:rhoeff}
    \end{align}
    Further, the inequality by \cite{weissman2003inequalities} gives that
    \begin{align}
        \| \hat P(\cdot |s,a) - P(\cdot | s,a)\|_1 \leq \sqrt{\frac{2|\Scal|}{n(s,a)} \ln \frac{2}{\delta}}.\label{eqn:weissman}
    \end{align}
    Combining both with a union bound over all $|\Scal \times \Acal|$ state-action pairs, we get that
    it is sufficient to choose $c = \sqrt{(1 + 4 |\Scal|)\ln (4|\Scal||\Acal| / \delta)} \geq \sqrt{2 |\Scal| \ln (4|\Scal||\Acal| / \delta)} + \sqrt{ \ln (4|\Scal||\Acal| / \delta)}/2$ to ensure that $\tilde c(s,a) \geq \|F_{\hat R(s,a)} - F_{R(s,a)} \|_{\infty}
     + \| \hat P(\cdot | s,a) -P(\cdot |s,a)\|_1$ allowing us to apply Lemma~\ref{appx:mdp:opt}. 
     
     However, we can improve this result by removing the polynomial dependency of $c$ on the number of states $|\Scal|$ as follows.
     Consider a fixed $(s,a)$ and denote $v(s',x) :=  \sum_{a' \in \Acal} \pi(a'|s')F_{\gamma Z(s',a') +  R(s,a)}(x)$ where we set $Z = Z_\pi$. Our goal is to derive a concentration bound on $\sum_{s' \in \Scal} \left(\hat P(s'|s,a) -P(s'|s,a)\right) v(s',x)$ that is tighter than the bound derived from
     $\| \hat P(\cdot |s,a) - P(\cdot | s,a)\|_1$. Note that $v$ is not a random quantity and hence $\sum_{s' \in \Scal} \left(\hat P(s'|s,a) -P(s'|s,a)\right) v(s',x)$ is a normalized sum of independent random variables for any $x$. To deal with the continuous variable $x$ which prevents us from applying a union bound over $x$ directly, we use a covering argument. Let $K \in \NN$ be arbitrary and consider the discretization set 
     \begin{align}
         \bar \Xcal = \{ x \in \RR ~ | ~ \exists k \in [K], ~ \exists s' \in \Scal, ~ \forall x' < x, ~ v(s', x') < k/K \leq v(s', x)\}.
     \end{align}
     Define $\bar v(s', x) = v(s', \max\{x' \in \bar \Xcal ~:~ x' \leq x\})$ as the discretization of $v$ at the discretization points in $\bar \Xcal$. This construction ensures that the discretization error is uniformly bounded by $1/K$, that is, $|\bar v(s,x) - v(s,x)| \leq 1/K$ holds for all $s \in \Scal$ and $x \in [\Vmin, \Vmax]$.
     Hence, we can bound for all $x \in [\Vmin, \Vmax]$
     \begin{align}
         & \sum_{s' \in \Scal} \left(\hat P(s'|s,a) -P(s'|s,a)\right) v(s',x)\\
         =& 
         \sum_{s' \in \Scal} \left(\hat P(s'|s,a) -P(s'|s,a)\right)\bar  v(s',x) +
         \sum_{s' \in \Scal} \left(\hat P(s'|s,a) -P(s'|s,a)\right) (v (s', x) - \bar  v(s',x))\\
         \circledmarked{1}{\leq} &
         \sqrt{\frac{1}{2n(s,a)} \ln \frac{|\bar \Xcal|}{\delta}}+
         \|\hat P(\cdot|s,a) -P(\cdot|s,a)\|_1 \|v (\cdot, x) - \bar  v(\cdot,x)\|_\infty\\
         \circledmarked{2}{\leq} & 
         \sqrt{\frac{1}{2n(s,a)} \ln \frac{|\bar \Xcal|}{\delta}} 
         + \frac{1}{K}\sqrt{\frac{2|\Scal|}{n(s,a)} \ln \frac{2}{\delta}}
     \end{align}
     where in \circledmarker{1}, we applied Hoeffding's inequality to the first term in combination with a union bound over $\bar \Xcal$ as $\bar v(\cdot, x)$ can only take $|\bar \Xcal|$ values in $\RR^{|\Scal|}$. The second term was bounded with H\"older's inequality and in \circledmarker{2} the concentration inequality from Eq.~\eqref{eqn:weissman} was used.
     Combining this bound with Eq.~\eqref{eqn:rhoeff} by applying a union bound over all states and actions, we get that picking
     \begin{align}
         \shift \geq \sqrt{\frac{1}{2}\ln \frac{3|\Scal||\Acal|}{\delta}}
     + \sqrt{\frac{1}{2}\ln \frac{3|\Scal||\Acal||\Xcal|}{\delta}}
     + \sqrt{\frac{2|\Scal|}{K^2}\ln \frac{6|\Scal||\Acal|}{\delta}}
     \end{align}
     is sufficient to apply Lemma~\ref{appx:mdp:opt}. Since $v(s', \cdot)$ is non-decreasing, the size of the discretization set is at most $|\bar \Xcal| \leq |\Scal|K$ and by picking $K = \sqrt{|\Scal|}$, we see that $\shift = O\left( \ln ( |\Scal||\Acal| / \delta)\right)$ is sufficient. 
     
\end{proof}

\begin{lemma}\label{appx:mdp:opt}
Let $Z, \tilde Z \in \Zcal$ be such that $F_{\tilde Z(s,a)} \leq F_{Z(s,a)}$ for all $(s,a)$. Assume finitely many states $|\Scal| < \infty$ and actions $|\Acal| < \infty$. Let $\hat{R}(s,a)$ and $\hat{P}(s,a)$ be the empirical reward distributions and transition probabilities.
Assume further that
\begin{align}
    \tilde \shift(s,a) \geq \|F_{\hat R(s,a)} - F_{R(s,a)} \|_{\infty}
     + \sum_{s' \in \Scal} \left(\hat P(s'|s,a) -P(s'|s,a)\right) \sum_{a' \in \Acal} \pi(a'|s')F_{\gamma Z(s',a') +  R(s,a)}(x) \label{eq:assum_opt_bonus}
\end{align}
where $\tilde \shift(s,a) = \frac{\shift}{\sqrt{n(s,a)}}$ is the shift in the optimism operator $O_\shift$.
Then $F_{O_\shift \hat \Tcal^{\pi}Z \tilde(s,a)} \leq F_{\Tcal^{\pi} Z(s,a)}$ for all $(s,a)$.
Note that it is sufficient to pick $\tilde \shift(s,a) \geq \|F_{\hat R(s,a)} - F_{R(s,a)} \|_{\infty} + \| \hat P(\cdot | s,a) -P(\cdot |s,a)\|_1$ to ensure the Assumption in Eq.~\eqref{eq:assum_opt_bonus}.
\end{lemma}
\begin{proof}
We start with some basic identities which follow directly from the definition of CDFs
\begin{align}
\label{eq:basicids}
    F_{P^\pi Z(s,a)}(x) 
    &=
    \sum_{s' \in \Scal, a' \in \Acal} P(s' | s, a) \pi(a' | s') F_{Z(s', a')}(x)\\
    F_{\gamma Z(s,a)}(x)  &= F_{Z(s,a)}(x / \gamma)\\
    F_{R(s,a) + Z(s,a)} (x) &=
    \int F_{Z(s,a)}(x - y) dF_{R(s,a)}(y) 
    = \int F_{R(s,a)}(x - y) dF_{Z(s,a)}(y)
\end{align}
where the integrals are Lebesgue-Stieltjes integrals and understood to be taken over $[V_{\min}, V_{\max}]$. We omit the limits in the following to unclutter notation. These identities allow us to derive expressions for $F_{\Tcal^{\pi} Z(s,a)}$ and $F_{O_\shift\hat{\Tcal}^{\pi} Z'(s,a)}$:
\begin{align}
    F_{\Tcal^{\pi} Z(s,a)} (x) &= F_{R(s,a)+\gamma P^{\pi}Z(s,a)} (x)\\ &= \int \sum_{s'\in \Scal, a'\in\Acal} P(s'|s,a)\pi(a'|s')F_{\gamma Z(s',a')}\left(x - y\right) dF_{R(s,a)}(y)\\
    &= \sum_{s'\in \Scal, a'\in\Acal} P(s'|s,a)\pi(a'|s') \int F_{\gamma Z(s',a')}\left(x - y\right) dF_{R(s,a)}(y)
    \\
    &= \sum_{s'\in \Scal, a'\in\Acal} P(s'|s,a)\pi(a'|s') F_{\gamma Z(s',a') + R(s,a)}(x)
    \label{eq:FPZ}\\
    F_{{O_\shift} \hat \Tcal^{\pi} \tilde Z(s,a)}(x) &= F_{{O_\shift} (\hat R(s,a)+\gamma \hat P^{\pi}\tilde Z(s,a))} (x) \\
    &= 0 \vee \left(\int \sum_{s'\in \Scal, a'\in\Acal} \hat P(s'|s,a)\pi(a'|s')F_{\gamma \tilde Z(s',a')}\left(x - y\right) dF_{\hat R(s,a)}(y) - \tilde \shift(s,a) \right)
    \\
    &= 0 \vee \left(\sum_{s'\in \Scal, a'\in\Acal} \hat P(s'|s,a)\pi(a'|s') F_{\gamma \tilde Z(s',a') + \hat R(s,a)}(x) - \tilde \shift(s,a) \right)
    \label{eq:OFPZ}
\end{align}
Here we exchanged the finite sum with the integral by linearity of integrals. Using these identities, we will show that
$F_{{O_\shift} \hat \Tcal^{\pi} \tilde Z(s,a)}(x)- F_{\Tcal^{\pi} Z(s,a)} (x) \leq 0$ for all $x$. Consider any fixed $x$ and first the case where the max in $F_{{O_\shift} \hat \Tcal^{\pi} \tilde Z(s,a)}(x)$ is attained by $0$. In this case $F_{{O_\shift} \hat \Tcal^{\pi} \tilde Z(s,a)}(x)- F_{\Tcal^{\pi} Z(s,a)} (x) =- F_{\Tcal^{\pi} Z(s,a)} (x) \leq 0$ because CDFs take values in $[0,1]$. For the second case, we combine~\eqref{eq:FPZ} and \eqref{eq:OFPZ} to write
\begin{align}
    & F_{{O_\shift} \hat \Tcal^{\pi} \tilde Z(s,a)}(x) - F_{\Tcal^{\pi} Z(s,a)} (x)\nonumber \\ 
    =&  \sum_{s' \in \Scal, a'\in \Acal} \hat{P}(s'|s,a) \pi(a'|s')F_{\gamma \tilde Z(s',a') +\hat R(s,a)}(x) - \tilde \shift(s,a)
    -\sum_{s' \in \Scal, a'\in \Acal} P(s'|s,a) \pi(a'|s') F_{\gamma Z(s',a') + R(s,a)}(x) \nonumber\\
    =&  -\tilde \shift(s,a) \nonumber\\
         & + \sum_{s' \in \Scal, a'\in \Acal} \hat P(s'|s,a) \pi(a'|s') 
      \left(F_{\gamma \tilde Z(s',a') + R(s,a)}(x) - F_{\gamma Z(s',a') + R(s,a)}(x)\right)
       \label{eq:FPZterm1}\\
      &+ \sum_{s' \in \Scal, a'\in \Acal} \hat P(s'|s,a) \pi(a'|s') 
      \left(F_{\gamma \tilde Z(s',a') + \hat R(s,a)}(x) - F_{\gamma \tilde Z(s',a') + R(s,a)}(x)\right)
      \label{eq:FPZterm2}\\
    &+ \sum_{s' \in \Scal, a'\in \Acal} \pi(a'|s') \left(\hat P(s'|s,a) -P(s'|s,a)\right) F_{\gamma Z(s',a') +  R(s,a)}(x).  \label{eq:FPZterm3}
    \end{align}
In the following, we consider each of the terms \eqref{eq:FPZterm1}--\eqref{eq:FPZterm3} separately. Let us start with \eqref{eq:FPZterm1} and bound
\begin{align}
    &F_{\gamma \tilde Z(s',a') + R(s,a)}(x) - F_{\gamma Z(s',a') + R(s,a)}(x)\nonumber\\
    &=\int \left[F_{\tilde Z(s,a)}\left(\frac{x - y}{\gamma}\right) - F_{Z(s,a)}\left(\frac{x - y}{\gamma}\right) \right]dF_{R(s,a)}(y)
    \leq \int 0 dF_{R(s,a)}(y) = 0
\end{align}
where the identity follows from the basic identities in Eq.~\eqref{eq:basicids} and the inequality from the assumption that $F_{\tilde Z(s,a)} \leq F_{Z(s,a)}$ for all $(s,a)$. Hence, the term in Eq.~\eqref{eq:FPZterm1} is always non-positive. Moving on to the term in Eq.~\eqref{eq:FPZterm2} which we bound with similar tools as
\begin{align}
    & F_{\gamma \tilde Z(s',a') + \hat R(s,a)}(x) - F_{\gamma \tilde Z(s',a') + R(s,a)}(x)\\
    =& \int \left[F_{\hat R(s,a)}\left(x - y\right) - F_{R(s,a)}\left(x - y\right) \right]dF_{\gamma \tilde Z(s,a)}(y)\\
    \leq&  \int \left|F_{\hat R(s,a)}\left(x - y\right) - F_{R(s,a)}\left(x - y\right) \right|dF_{\gamma \tilde Z(s,a)}(y)\\
    \leq & \sup_{z}\left|F_{\hat R(s,a)}\left(z\right) - F_{R(s,a)}\left(z\right) \right|\int  dF_{\gamma \tilde Z(s,a)}(y) 
    = \left\|F_{\hat R(s,a)} - F_{R(s,a)} \right\|_{\infty}.
\end{align}
This yields that Eq.~\eqref{eq:FPZterm2} is bounded by this as well
\begin{align}
    \sum_{s' \in \Scal, a'\in \Acal} \hat P(s'|s,a) \pi(a'|s') \left\|F_{\hat R(s,a)} - F_{R(s,a)} \right\|_{\infty} = \left\|F_{\hat R(s,a)} - F_{R(s,a)} \right\|_{\infty}. \label{eq:FPZterm2_bnd}
\end{align}
Finally, the last term from Eq.~\eqref{eq:FPZterm3} can be bounded as follows
\begin{align}
&\sum_{s' \in \Scal}  \left(\hat P(s'|s,a) -P(s'|s,a)\right) \sum_{a'\in \Acal} \pi(a'|s') F_{\gamma Z(s',a') +  R(s,a)}(x)\\
\leq & \| \hat P(\cdot | s,a) -P(\cdot |s,a)\|_1 \left\| \sum_{a'\in \Acal} \pi(a'|\cdot) F_{\gamma Z(s',a') + R(s,a)}(x) \right\|_\infty\\
\leq & \| \hat P(\cdot | s,a) -P(\cdot |s,a)\|_1 \left\| \sum_{a'\in \Acal} \pi(a'|\cdot) \times 1 \right\|_\infty
\leq  \| \hat P(\cdot | s,a) -P(\cdot |s,a)\|_1 \label{eq:FPZterm3_bnd}
\end{align}
Combining the individual bounds for each of the terms \eqref{eq:FPZterm1}--\eqref{eq:FPZterm3}, we end up with 
\begin{align}
     & F_{{O_\shift} \hat \Tcal^{\pi} \tilde Z(s,a)}(x) - F_{\Tcal^{\pi} Z(s,a)} (x)\\
     \leq & -\tilde \shift(s,a)
     + \|F_{\hat R(s,a)} - F_{R(s,a)} \|_{\infty}
     + \sum_{s' \in \Scal}  \left(\hat P(s'|s,a) -P(s'|s,a)\right) \sum_{a'\in \Acal} \pi(a'|s') F_{\gamma Z(s',a') +  R(s,a)}(x)\\
     \leq & -\tilde \shift(s,a)
     + \|F_{\hat R(s,a)} - F_{R(s,a)} \|_{\infty}
     + \| \hat P(\cdot | s,a) -P(\cdot |s,a)\|_1,
\end{align}
which is non-positive as long as $\tilde \shift(s,a) \geq \|F_{\hat R(s,a)} - F_{R(s,a)}\|_{\infty}
     + \| \hat P(\cdot | s,a) -P(\cdot |s,a)\|_1$ which completes the proof.
\end{proof}

\subsection{Technical lemmas}
\begin{lemma}\label{lem:cvarcdf}
    Let $F$ and $G$ be the CDFs of two non-negative random variables and let $\nu_F$, $\nu_G$ be a maximizing value of $\nu$ in the definition of $\cvar_{\alpha}(F)$ and $\cvar_{\alpha}(G)$ respectively. Then:
    \begin{align}
        |\cvar_{\alpha}(F) - \cvar_{\alpha}(G)| \leq & \frac 1 \alpha \int_{0}^{\max\{F^{-1}(\alpha), G^{-1}(\alpha)\}} |G(y) - F(y)| dy\\
        \leq & \frac{\max\{F^{-1}(\alpha), G^{-1}(\alpha)\}}{\alpha} \sup_{x}|F(x) - G(x)|
    \end{align}
\end{lemma}
\begin{proof}
   Assume w.l.o.g. that $\cvar_{\alpha}(F) - \cvar_{\alpha}(G) \geq 0$. Denote by $\nu_F$ any maximizing value of $\nu$ in the definition of  $\cvar_{\alpha}(F)$. By Lemma~4.2 and Equation~{(4.9)} in \cite{acerbi2002coherence}, a possible value of $\nu_F$ is $F^{-1}(\alpha)$. Then we can write the differences in CVaR as
    \begin{align}
     \cvar_{\alpha}(F) - \cvar_{\alpha}(G)
     \leq & \nu_F - \alpha^{-1}\EE_F[(\nu_F - X)^+] -(\nu_F - \alpha^{-1}\EE_G[(\nu_F - X)^+])\\
          = & \frac 1 \alpha (\EE_G[(\nu_F - X)^+] - \EE_F[(\nu_F - X)^+]).\label{eqn:diffE1}
    \end{align}
     Using Lemma~\ref{lem:cvar_expression} in Equation~\eqref{eqn:diffE1} gives
    \begin{align}
    \cvar_{\alpha}(F) - \cvar_{\alpha}(G)
    \leq & \frac 1 \alpha \left( \int_{0}^{\nu_F} G(y) dy - \int_{0}^{\nu_F} F(y) dy\right)\\
    \leq & \frac 1 \alpha \int_{0}^{\nu_F} |G(y) - F(y)| dy
    \leq \frac{\nu_F}{\alpha} \sup_{y} |F(y) - G(y)|.
    \end{align}
    We can in full analogy upper-bound $\cvar_{\alpha}(G) - \cvar_{\alpha}(F)$ and arrive at the desired statement.
\end{proof}

\begin{lemma}\label{lem:cvar_expression}
Let $F$ be a CDF of a bounded non-negative random variable and $\nu \in \RR$ be arbitrary. Then
$\EE_F[(\nu - X)^+] = \int_{0}^{\nu} F(y) dy$. Hence, 
one can write the conditional value at risk of a variable $X \sim F$ for any CDF $F$ with $F(0) = 0$  as
\begin{align}
    \cvar_{\alpha}(F) = \sup_{\nu} \left\{ \frac 1 \alpha \int_{0}^{\nu} (\alpha - F(y)) dy\right\}.
\end{align}
\end{lemma}
\begin{proof}
We rewrite $\EE_F[(\nu - X)^+]$ as follows
    \begin{align}
        \EE_F[(\nu - X)^+] 
        =& \EE_F[(\nu - X)\one\{ X \leq \nu\}]
        = \nu F(\nu) - \EE_F[X \one\{ X \leq \nu\}] \\
        \circledmarked{1}{=}&
        \nu F(\nu) - \EE_F\left[ \one\{ X \leq \nu\}
        \int_{0}^{\infty} \one\{ X > y\} dy
        \right]\\
                \circledmarked{2}{=}&
        \nu F(\nu) - \int_{0}^{\infty}
        \PP_F\left[ y < X \leq \nu\right]dy
        \\
            =&
        \nu F(\nu) - \int_{0}^{\nu} (F(\nu) - F(y))dy
        = \int_{0}^{\nu} F(y) dy
    \end{align}
    where $\circledmarker{1}$ follows from $a = \int_0^a dx = \int_0^{\infty} \one\{a > x \} dx$ which holds for any $a \geq 0$ and $\circledmarker{2}$ uses Tonelli's theorem to exchange the two integrals. Plugging this identity into
    \begin{align}
        \nu - \frac 1 \alpha \EE_F[(\nu - X)^+] 
        = \frac 1 \alpha \left(\nu \alpha -  \int_{0}^{\nu} F(y) dy\right)
        = \frac 1 \alpha \int_{0}^{\nu} (\alpha - F(y)) dy
    \end{align} 
    and taking the sup over $\nu$ gives the desired result.
\end{proof}

\begin{lemma}\label{lem:stochdom_implies_cvaropt}
    Let $G$ and $F$ be CDFs of non-negative random variables so that $\forall x \geq 0: \quad F(x) \geq G(x)$. Then for any $\alpha \in [0, 1]$, we have $\cvar_{\alpha}(F) \leq \cvar_{\alpha}(G)$.
\end{lemma}
\begin{proof}Consider now the following difference
    \begin{align}
        \frac 1 \alpha \int_{0}^{\nu} (\alpha - G(y)) dy - \frac 1 \alpha \int_{0}^{\nu} (\alpha - F(y)) dy
        =         \frac 1 \alpha \int_{0}^{\nu} (F(y) - G(y)) dy \geq 0.
    \end{align}
    By Lemma~\ref{lem:cvar_expression}, we have that
    \begin{align}
        &\cvar_\alpha(G) - \cvar_\alpha(F) \\
        &=
        \sup_{\nu} \left\{\frac 1 \alpha \int_{0}^{\nu} (\alpha - G(y)) dy\right\} - \sup_{\nu} \left\{\frac 1 \alpha \int_{0}^{\nu} (\alpha - F(y)) dy\right\}.
    \end{align}
    Let $\nu_F$ denote a value of $\nu$ that achieves the supremum in $\frac 1 \alpha \int_{0}^{\nu} (\alpha - F(y)) dy$ (which exists by Lemma~4.2 and Equation~{(4.9)} in \cite{acerbi2002coherence}). Then
    \begin{align}
        \cvar_\alpha(G) - \cvar_\alpha(F) \geq 
                \frac 1 \alpha \int_{0}^{\nu_F} (\alpha - G(y)) dy
                - \frac 1 \alpha \int_{0}^{\nu_F} (\alpha - F(y)) dy \geq 0.
    \end{align}
\end{proof}

\begin{lemma}
    Consider a sequence of CDFs $\{F_{n}\}$ on $x \in [\Vmin, \Vmax]$ with $\Vmin \geq 0$ that converges in $\ell_2$ distance to $F_{O}$ as $n \rightarrow \infty$. Then $\cvar_{\alpha}(F_{n}) 
\rightarrow \cvar_{\alpha}(F_{O})$ as $n \rightarrow \infty$.
\end{lemma}
\begin{proof}
    Consider the sequence of Wasserstein $d_1$ distance between the CDFs:
    \begin{align}
        d_1 \left(F_{n} , F_{Z_O}\right) &= \int_{\Vmin}^{\Vmax}|F_{n}(x) - F_{O}(x)| dx\\ 
        &\leq \left( \int_{\Vmin}^{\Vmax}|F_{n}(x) - F_{O}(x)|^2 dx \right)^{1/2} \left( \int_{\Vmin}^{\Vmax} 1 dx \right)^{1/2}\\
        & = \sqrt{\Vmax - \Vmin} \ell_2(F_{n} , F_{O})
    \end{align}
    Where the second line follows by H\"older's inequality. The right hand side goes to $0$ as $n \rightarrow \infty$, which implies convergence in Wasserstein $d_1$ distance ($p$=1). Finally, using Lemma~\ref{lem:cvarcdf}, convergence of CVaR follows.
\end{proof}

\end{document}